\documentclass[lettersize,journal]{IEEEtran}
\usepackage{amsmath,amsfonts}
\usepackage{array}
\usepackage[caption=false,font=normalsize,labelfont=sf,textfont=sf]{subfig}
\usepackage{textcomp}
\usepackage{stfloats}
\usepackage{url}
\usepackage{verbatim}
\usepackage{graphicx}
\usepackage{cite}
\usepackage{graphicx}
\usepackage{amsmath}
\usepackage{amssymb}
\DeclareMathOperator{\E}{\mathbb{E}}
\usepackage{booktabs}
\usepackage{multirow}
\usepackage{subfig}
\usepackage[export]{adjustbox}

\usepackage{makecell}
\usepackage{multirow}  % multirow 패키지 사용
\usepackage{pifont}
\usepackage{xcolor}
\newcommand{\cmark}{\textcolor{green!80!black}{\ding{51}}}
\newcommand{\xmark}{\textcolor{red}{\ding{55}}}

\usepackage[linesnumbered, ruled, lined]{algorithm2e} % to write algorithm

\SetCommentSty{mycommfont}
\usepackage{physics}
\usepackage{orcidlink}
\usepackage{kotex}

\usepackage[mathlines,switch]{lineno} % peer review, line number

\definecolor{mypink}{rgb}{0,0,0} % for clean version
\definecolor{myblue}{rgb}{0,0,0} % for clean version
\definecolor{mygreen}{rgb}{0,0,0} % for clean version
\definecolor{minormyblue}{rgb}{0,0,0} % for clean version
\definecolor{minormygreen}{rgb}{0,0,0} % for clean version
\newcommand{\news}[1]{\textcolor{mypink}{#1}} %mypink
\newcommand{\firstR}[1]{\textcolor{myblue}{#1}}
\newcommand{\secondR}[1]{\textcolor{mygreen}{#1}}

\newcommand{\minorfirstR}[1]{\textcolor{minormyblue}{#1}}
\newcommand{\minorsecondR}[1]{\textcolor{minormygreen}{#1}}

\begin{document}

% \title{Q-HyViT: Post-Training Quantization for Hybrid Vision Transformers \\with Bridge Block Reconstruction}
\title{Q-HyViT: Post-Training Quantization of Hybrid Vision Transformers with Bridge Block Reconstruction \minorfirstR{for IoT Systems}}

 \author{Jemin Lee\orcidlink{0000-0002-9332-3508},
 Yongin Kwon\orcidlink{0000-0003-2973-246X},
 Sihyeong Park\orcidlink{0000-0001-8244-4817},
 Misun Yu\orcidlink{0000-0001-7319-1053},
 Jeman Park\orcidlink{0009-0002-9524-0738},
  and Hwanjun Song\orcidlink{0000-0002-1105-0818}
        % <-this % stops a space
\thanks{This work was supported by Institute of Information \& communications
Technology Planning \& Evaluation (IITP) grant funded by the Korea
government(MSIT) (No.RS-2023-00277060, Development of open edge AI SoC hardware and software platform)
(Corresponding author: Hwanjun Song)}
\thanks{Jemin Lee, Yongin Kwon, Misun Yu, and Jeman Park are with the Artificial Intelligence Computing Research Laboratory, Electronics and Telecommunications Research Institute(ETRI), Republic of Korea (e-mail:\{leejaymin,yongin.kwon,msyu,jeman\}@etri.re.kr)}

\thanks{Sihyeong Park is with the SoC Platform Research Center, Korea Electronics Technology Institute(KETI), Republic of Korea (e-mail:sihyeong@keti.re.kr)}

\thanks{Hwanjun Song is with the Department of Industrial and Systems Engineering, KAIST, South Korea (e-mail:songhwanjun@kaist.ac.kr)}}
% <-this % stops a space

\markboth{Journal of \LaTeX\ Class Files,~Vol.~X, No.~X, May~2024}%
{Shell \MakeLowercase{\textit{et al.}}: A Sample Article Using IEEEtran.cls for IEEE Journals}

\IEEEpubid{0000--0000/00\$00.00~\copyright~2023 IEEE}

\maketitle

\modulolinenumbers[3] % line number
%\linenumbers % line number

\begin{abstract}
Recently, vision transformers (ViTs) have superseded convolutional neural networks in numerous applications, including classification, detection, and segmentation. However, the high computational requirements of ViTs hinder their widespread implementation. To address this issue, researchers have proposed efficient hybrid transformer architectures that combine convolutional and transformer layers with optimized attention computation of linear complexity. Additionally, post-training quantization has been proposed as a means of mitigating computational demands. For mobile devices, achieving optimal acceleration for ViTs necessitates the strategic integration of quantization techniques and efficient hybrid transformer structures. However, no prior investigation has applied quantization to efficient hybrid transformers.
In this paper, we discover that applying existing post-training quantization (PTQ) methods for ViTs to efficient hybrid transformers leads to a drastic accuracy drop, attributed to the four following challenges:  (i) highly dynamic ranges, (ii) zero-point overflow, (iii) diverse normalization, and (iv) limited model parameters ($<$5M). 
To overcome these challenges, we propose a new post-training quantization method, which is the first to quantize efficient hybrid ViTs (MobileViTv1, MobileViTv2, Mobile-Former, EfficientFormerV1, EfficientFormerV2). We achieve a significant improvement of \firstR{17.73\% for 8-bit and 29.75\% for 6-bit on average, respectively, compared with existing PTQ methods (EasyQuant, FQ-ViT, PTQ4ViT, and RepQ-ViT)}. We plan to release our code at \textcolor{blue}{\url{https://gitlab.com/ones-ai/q-hyvit}}.
\end{abstract}


\section{Introduction}
\IEEEPARstart{R}{cent} 
\label{sec:intro}
\secondR{advancements in quantization method have markedly enhanced the integration of federated learning and deep neural networks within the Internet of Things (IoT) domain~\cite{chen2023federated,ji2022fedqnn,liu2023quasyncfl}, facilitating a significant leap towards more efficient computation and communication. These advancements enable IoT devices to engage in distributed, privacy-preserving machine learning, transforming them into highly intelligent systems capable of real-time data analysis and autonomous decision-making. 
Through optimized hardware accelerators and communication-efficient protocols, quantization paves the way for the deployment of advanced AI models on edge devices, heralding a new era of intelligent, efficient, and autonomous IoT applications.}

Meanwhile, thanks to self-attention that captures the global representation and shows better generalization with a low inductive bias, vision transformers (ViTs) have substituted convolutional neural networks (CNNs) in numerous applications, such as image classification, object detection, and instance segmentation~\cite{han2022survey, khan2022transformers}. 
Despite the great success of ViT, the high computational requirement of ViTs still remains a significant impediment to their widespread implementation.

To democratize the use of ViT on resource-constrained devices, researchers have proposed a \emph{hybrid} vision transformer architectures, which combine convolutional and transformer layers, such as MobileViTv1~\cite{mehta2021mobilevit}. They have also optimized attention computation to achieve linear complexity, such as MobileViTv2~\cite{mehta2022separable}. 
Additionally, quantization techniques have been adopted for efficient architecture design, achieving model compression by reducing the precision of float values.
Typically, the quantization techniques are categorized into two types: {quantization-aware training} (QAT) and {post-training quantization} (PTQ). While QAT offers advantages in preserving accuracy compared with PTQ, its adoption has been restricted due to privacy concerns, the resource-intensive and time-consuming nature of the re-training process, and the requisite expertise for hyperparameter tuning in architecture development~\cite{kris_whitepaper2018,steven2020,PACT2018,zhang2018lq,jung2019learning,zhou2016dorefa,jacob2018,songhan2016}. 

\IEEEpubidadjcol
In practical setup, PTQ methods have been more commonly employed due to their high applicability~\cite{kris_whitepaper2018, jiang2021automated, banner_neurips2019, choukroun2019low, zhao2019improving, lee2018quantization, goncharenko2019fast, migacz20178, wu2020integer}. 
PTQ enables the calibration of pre-trained models, utilizing only a small unlabeled dataset. 
PTQ for CNN models has been studied extensively, and recently, there has been a notable surge in interest regarding PTQ for ViTs. 
PTQ for ViTs shows its ability to maintain the accuracy of quantized models, effectively addressing varied activation ranges resulting from a non-linear function. However, these studies have solely focused on {canonical} transformer architectures, such as ViT~\cite{dosovitskiy2020image}, DeiT~\cite{touvron2021training}, and Swin Transformers~\cite{liu2021swin}.

\iffalse
\begin{figure}[t]
	\centering
	\includegraphics[width=1\columnwidth]{./figures/overview2}
	%\includegraphics[width=1\columnwidth]{./fig/overview}
	\caption{Overall quantization process of Q-HyViT on the representative structure of hybrid vision transformers, including local, global, and bridge representation.}
	\label{fig:overview}
\end{figure}
\fi

\begin{figure}[t]
	\centering
	\includegraphics[width=1\columnwidth]{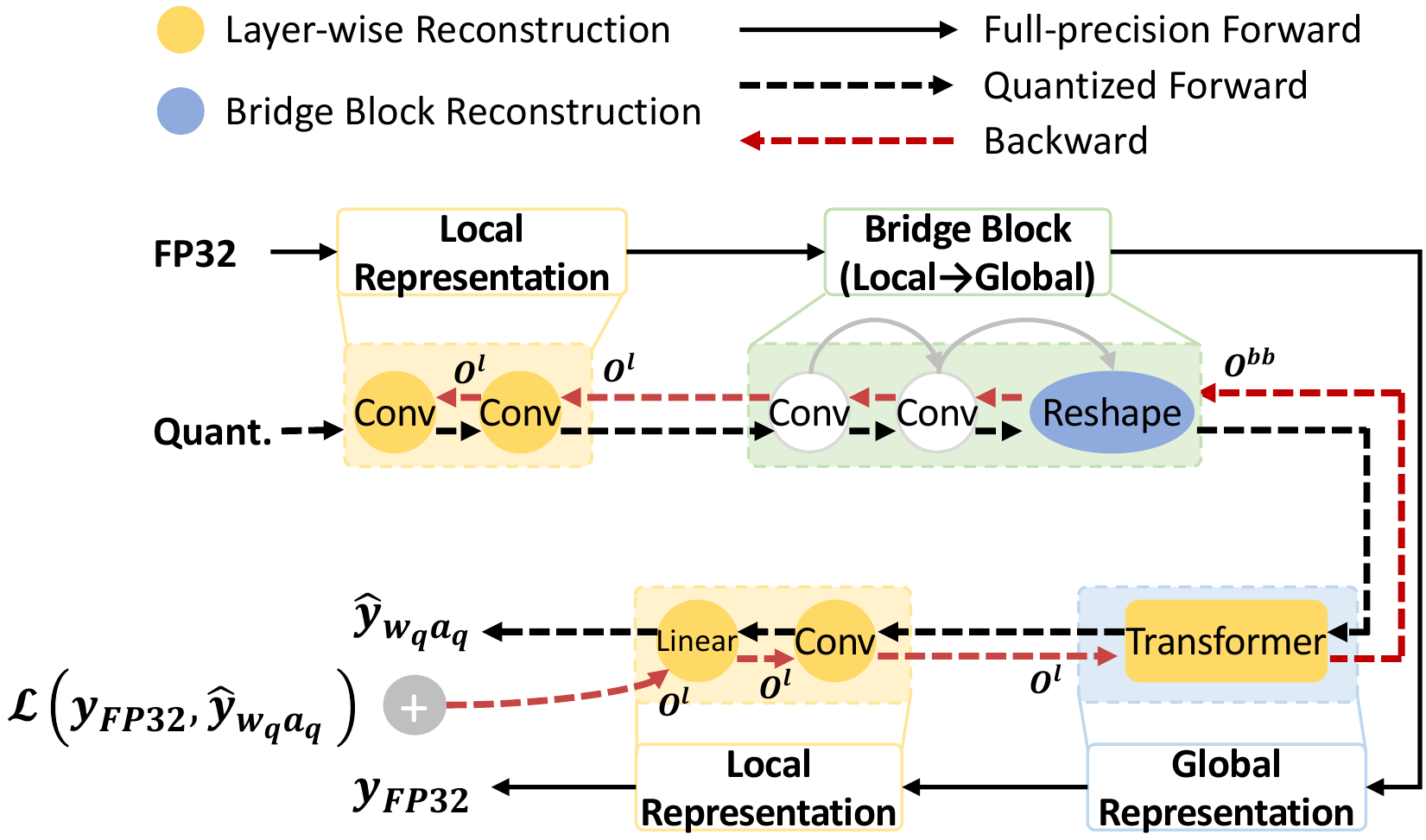}
	\caption{Overall quantization process of Q-HyViT on the representative structure of hybrid vision transformers, including local, global, and bridge representation.}
	\label{fig:overview}
\end{figure}

For mobile devices, achieving optimal acceleration for ViTs necessitates the integration of quantization techniques and efficient transformer structures. 
However, there has been no prior exploration into applying quantization to efficient {hybrid} transformers. 
%Our study addresses this gap by directly applying existing post-training quantization techniques to hybrid vision transformers, which poses the following challenges,
While the existing PTQ can be directly applied to hybrid ViTs, this process is non-trivial due to the four key differences distinguishing them from the canonical ViTs:
%To investigate the potential of directly applying the existing PTQ to hybrid ViT, we identify four key issues that hinder existing PTQ: 
(i) \emph{the highly dynamic activation range}, which complicates accuracy preservation using existing methodologies for pure ViTs; 
(ii) \emph{the existence of bridge blocks}, which serves as connectors between convolution and transformer layers, introducing a disparity between local and global representations and the issue of zero-point overflow; 
(iii) \emph{the diverse types of normalization techniques}, which are used in hybrid vision transformers; and 
(iv) \emph{the small-sized models with less than 5 million parameters}, which lead to a substantial loss of robustness during quantization due to their limited number of parameters and minimal residual connections.
Therefore, it is imperative to simultaneously adjust the granularity and scheme of both bridge and non-bridge layers, while also identifying the optimal scaling factors for quantization.

%We propose a new post-training quantization technique that overcomes these challenges and is the first work to quantize hybrid vision transformers (MobileViTv1 and MobileViTv2) with a significant margin compared to existing techniques. 
%Our method reduces the accuracy drop concerning MobileViTv1 and MobileViTv2. Additionally, we will soon release our code as open-source software.

In this paper, we propose \textbf{Q-HyViT} in Figure~\ref{fig:overview}, a tailored quantization approach for hybrid vision transformers aimed at minimizing quantization error.
Q-HyViT integrates a novel \emph{hybrid reconstruction} error minimization technique, which can determine the optimal scale factors, granularity (channel-wise or layer-wise), and quantization scheme (symmetric or asymmetric). Here, the reconstruction objective of each layer differs based on whether it is part of the bridge block\footnote{The term ``bridge block'' literally refers to the transition part between convolution and transformer blocks. The precise constitution of this block slightly differs among hybrid ViT architectures.} or not.
To implement this technique in an integrated manner, we reuse the second-order term-based reconstruction error minimization method and extend it to incorporate the bridge block.
To our knowledge, this is the \emph{first} work that identifies the challenges of quantization for hybrid vision transformers and proposes a unified method to mitigate their errors in post-training quantization. 

% how the algorithm works

% describe evaluation results
We conduct comprehensive experiments to compare Q-HyViT with existing open-source quantization methods, namely EasyQuant~\cite{wu2020easyquant}, FQ-ViT~\cite{lin2022fq}, PTQ4ViT~\cite{yuan2022ptq4vit}, \firstR{and RepQ-ViT~\cite{li2023repq}}, on the same hybrid ViTs. %The results demonstrate the preservation of accuracy degradation caused by quantization in Q-HyViT. 
The experiments use hybrid ViTs including MobileViTv1~\cite{mehta2021mobilevit}, MobileViTv2~\cite{mehta2022separable}, Mobile-Former~\cite{chen2022mobile}, EfficientFormerV1~\cite{li2022efficientformer}, and EfficientFormerV2~\cite{li2022rethinking}, which are representative variants of efficient ViTs.

The results demonstrate that our Q-HyViT performs considerably well across five types of hybrid ViTs and outperforms existing PTQ methods for pure ViTs (EasyQuant, PTQ4ViT, \firstR{and RepQ-ViT}) by a large margin (up to an average improvement of 17.73\% for 8-bit and 29.75\% for 6-bit). 
Particularly, we highlight that in \emph{full} quantization setup where quantizing non-linear operations (softmax and normalization) is essential, Q-HyViT achieves state-of-the-art accuracy (an average improvement of 43.63\% for 8-bit compared with FQ-ViT) on hybrid ViTs. 

%- direct implementation
%- comparison in quantization
%- comparison in a fully quantized model 
%- ablation study

% ptq4vit
% Experiments show the quantized vision transformers (ViT, DeiT, and Swin) achieve near-lossless prediction accu- racy (less than 0.5% drop at 8-bit quantization) on the ImageNet classification task.
% apq
% Our APQ-ViT revisits the process of post-training quantization
%for vision transformer and presents novel insights. Comprehensive experiments on the large-scale computer vision tasks (image classifi- cation [7] and object detection [26]) demonstrate that our APQ-ViT performs remarkably well across various transformer architectures such as ViT [9], DeiT [40], and Swin Transformer [28], and sur- passes the existing methods by convincing margins, especially in lower bit-width settings (e.g., averagely up to 5.17% improvement for classification and 24.43% for detection on W4A4). We highlight that our APQ-ViT scheme achieves state-of-the-art accuracy per- formance on various bit-width settings, and enjoys versatility on diverse architectures and vision tasks.

Our primary contributions are summarized below:
\begin{itemize}

\item We discover that quantization of hybrid ViTs presents four unique challenges: (i) the presence of highly dynamic activation ranges, (ii) zero-point overflow in the bridge block, (iii) diverse normalization, and (iv) a parameter count of less than 5 million.

\item We propose a unified method called Q-HyViT, which is based on Hessian to adjust the granularity and scheme for bridge and non-bridge layers, while also determining the optimal scaling factors for quantization.

% \item For comparison, we re-implement the existing PTQ methods enabling them to be applied to hybrid ViTs. We then evaluate Q-HyViT for large-scale image classification using five variants of efficient hybrid ViTs.
% \item Our experimental results demonstrate the effectiveness of Q-HyViT in preserving the accuracy of quantized hybrid ViTs, as it outperforms the state-of-the-art PTQ methods, FQ-ViT and PTQ4ViT.

\item \firstR{We extend the existing PTQ methods to accommodate the five variants of efficient hybrid ViTs and directly compare them with our proposed Q-HyViT framework. Our experimental results reveal that Q-HyViT significantly outperforms state-of-the-art PTQ methods, including FQ-ViT, PTQ4ViT, and RepQ-ViT, in maintaining the accuracy of quantized hybrid ViTs.}

\end{itemize}

% 5m under 의 모델을 시도했다. (cvpr18논문 논리)
% PTQ 이야기도 넣기 
\iffalse
CNN 방법론 적용시 -> 문제점
Transformer 방법론 적용시 -> 문제점
표를 만들어서 각 케이스별로 해결법을 구조적 나열 (기존 방법들의 조합 or a little bit new)
새로운 메소드를 제안x -> 기존의 기법들을 hybrid 생태게에 맞게 동작 시킨다. 간단하게 동작 
\fi
% 5m under 의 모델을 시도했다. (cvpr18논문 논리)

\section{Related Work}
Model architecture design and quantization techniques have received substantial attention in the context of efficient AI. 
We provide an overview of prior research endeavors on designing efficient ViT architectures and efficient quantization methods.
%, followed by a comparison of our novelty against the existing studies.

\subsection{Efficient Computer Vision Architecture}
To avoid heavy computation in CNNs, standard convolutions have been replaced with separable convolutions~\cite{chollet2017xception}.
Separable convolutions have been widely used when designing light-weight CNNs including MobileNets~\cite{howard2017mobilenets,sandler2018mobilenetv2,howard2019searching}, ShuffleNetv2~\cite{ma2018shufflenet}, and MNASNet~\cite{tan2019mnasnet}, reducing the computational complexity of CNN operations.
Despite the prevalence of these models, a major drawback of them is that they are spatially local and have higher inductive bias.

%-> global and local representation (shape filter, texture filter) 
To expand model capacity while minimizing inductive bias, transformers~\cite{dosovitskiy2020image} are employed for computer vision, where pure transformers are directly applied to process image patches as a sequence. 
Dosovitskiy~\textit{et~al.}~\cite{dosovitskiy2020image} showed that it performs better than recent CNN-based architectures on multiple image recognition benchmarks. 
Moreover, Touvron~\textit{et~al.}~\cite{touvron2021training} introduced a teacher-student strategy tailored for transformers, resulting in competitive convolution-free transformers trained solely on ImageNet data. 

% hybrid vision transformer (v1, v2)
Even though pure ViT models achieve performance competitive to CNNs, the majority of them are computationally intensive.
Recently, MobileViTv1~\cite{mehta2021mobilevit}, MobileViTv2~\cite{mehta2022separable}, EfficientFormerV1~\cite{li2022efficientformer}, EfficientFormerV2~\cite{li2022rethinking}, and Mobile-Former~\cite{chen2022mobile} have been proposed for lightweight ViTs.
Such hybrid ViTs show accuracy higher than light-weight CNNs (MobileNet series) with a comparable number of model parameters by incorporating a fusion of convolution and transformer layers.
%Furthermore, Mehta~\textit{et~al.} proposed MobileViTv1 that uses separable self-attention to reduce attention complexity ($O(n^2)$ to $O(n)$)~\cite{mehta2022separable}. 
%Considering these active studies on efficient model architecture design, the hybrid vision transformer will continue to replace CNNs in many areas, and its usefulness will be significant in terms of optimizing the model simultaneously.

\subsection{Model Quantization}
Quantization methods are categorized into two types: quantization-aware training (QAT) and post-training quantization (PTQ).
Although QAT methods have successfully mitigated the accuracy degradation of quantized models by mapping from high-bit to low-bit precision~\cite{kris_whitepaper2018,steven2020,PACT2018,zhang2018lq,jung2019learning,zhou2016dorefa,jacob2018,songhan2016}, their widespread adoption has been hindered due to dependencies on re-training, the necessity of a complete dataset, and sophisticated hyper-parameter tuning.

Post-training quantization methods, which convert high-precision representation bits to low-precision bits without requiring re-training steps, have been extensively studied and widely adopted in practical scenarios~\cite{jiang2021automated,choukroun2019low,nagel2019data,zhao2019improving,lee2018quantization,goncharenko2019fast,meller19a,migacz20178,wu2020integer}. 
PTQ helps in the rapid deployment of CNN models on resource-constrained devices by addressing time-consuming and data privacy concerns. However, PTQ leads to significant accuracy degradation, particularly in low-precision representations, and prior research has mainly focused on \texttt{int8} quantization. 
As an effort to preserve the performance of a full-precision model, recent PTQ works~\cite{adaround20,brecq21,hubara2021accurate,wei2022qdrop,wangleveraging} have suggested to reconstruction error minimization of each layer or block by adjusting the magnitude of weight rounding and searching optimal scaling factors. 

Recently, quantization for ViTs has been studied~\cite{liu2021post,lin2022fq,yuan2022ptq4vit,apq22,liu2023noisyquant, li2023repq}.
In efforts to minimize quantization errors, these research endeavors have taken into account the unique structure of ViTs, such as multi-head attention and layer normalization. 
However, they do not account for the distinctive characteristics of hybrid ViTs -- the presence of bridge blocks and the utilization of diverse normalization techniques.

\section{Preliminary}
% show problematic distribution (observation)

\subsection{Hybrid Vision Transformers and Bridge Blocks}
\firstR{To address the inefficiencies of canonical ViTs, hybrid ViTs have been proposed to combine convolutional operations with transformers. These aim to reduce model size without compromising on accuracy.}

\subsubsection{Variants of Hybrid Vision Transformer} From a broader perspective, recent hybrid vision transformers can be classified into three categories, namely MobileViTseries~\cite{mehta2021mobilevit,mehta2022separable}, Mobile-Former~\cite{chen2022mobile}, and EfficientFormer series~\cite{li2022efficientformer,li2022rethinking}.

\firstR{
\textbf{MobileViT series:} The fundamental principle underneath MobileViT's design philosophy focuses on bringing together CNNs and ViTs in order to attain an optimal balance between efficiency and performance in the field of mobile vision tasks. By integrating transformer blocks within a CNN framework, MobileViTv1 skillfully processes both local and global information, outperforming traditional models in parameter efficiency. The subsequent version, MobileViTv2, advances this approach by transforming the attention map from a quadratic $O(N^2)$ to linear $O(N)$ complexity, further enhancing computational efficiency.
}

\firstR{
\textbf{Mobile-Former:} The model excels by using MobileNet and Transformers together, bridging them with a bi-directional connector. This unique setup enables an effective exchange of local and global features, optimizing the model's performance on mobile devices without compromising efficiency.}

\firstR{
\textbf{EfficientFormer series:} This architecture differs from the use of MobileNet blocks and instead uses a convolutional stem followed by a pure transformer architecture specifically optimized for mobile usage. EfficientFormerV1 establishes a new benchmark by effectively tackling the common inefficiencies seen in ViTs, including channel size reduction and spatial downsampling. The EfficientFormerV2 model expands upon the existing framework by including a unified Feed Forward Network that incorporates depth-wise convolutions, a more condensed and deeper architecture, and a dual-path attention downsampling method.}

\subsubsection{Bridge Blocks} The Bridge Block varies in its exact location depending on each hybrid vision model, but fundamentally consists of operators that adjust dimensions to facilitate transitions between local and global representations. Specifically, for MobileViTv1 and MobileViTv2, it implies convolution and reshape operators to align the input dimensions of the Transformer. In the case of Mobile-Former, it refers to operators within the Mobile-Former Block that transition bi-directionally between local and global representations.
For Efficient-FormerV1, these are operators that convert meta-blocks from 4D to 3D. In the case of Efficient-FormerV2, the bridge block consists of local and global transition operators present in the 3rd and 4th stages.

\begin{table*}[t]
\centering
\caption{Definition and precise configuration of the Bridge Block within each hybrid vision transformer}
\label{table:bridge_block}
\resizebox{\textwidth}{!}{
\begin{tabular}{lllc}
\toprule
\multicolumn{1}{c}{\textbf{Model}} &
\multicolumn{1}{c}{\textbf{Description}} &
\multicolumn{1}{c}{\textbf{Detailed Operators}} &
\multicolumn{1}{c}{\textbf{\makecell{\# of Bridge \\ Blocks}}} 
\\   \midrule
\midrule
\multirow{4}{*}{MobileViTv1} & \multirow{4}{*}{\makecell{Convolution and reshape operators \\ to align the input dimensions \\ of the Transformer
}} & \multirow{4}{*}{\texttt{\makecell{stages.2.1.convkxk.conv -> stages.2.1.conv1x1 \\ 
stages.3.1.convkxk.conv -> stages.3.1.conv1x1 \\
stages.4.1.convkxk.conv -> stages.4.1.conv1x1}}}  & \multirow{4}{*}{3} \\ 
  &  &  & \\ 
  &  &  &  \\
 &  &  &  \\  

\midrule
\multirow{7}{*}{MobileViTv2} & \multirow{7}{*}{\makecell{Convolution and reshape operators \\ to align the input dimensions \\ of the Transformer}} & \multirow{7}{*}{\texttt{\makecell{stages.2.1.convkxk.conv ->
stages.2.1.conv1x1 \\
stages.3.1.convkxk.conv ->
stages.3.1.conv1x1 \\
stages.4.1.convkxk.conv ->
stages.4.1.conv1x1}}} & \multirow{7}{*}{3} \\ %68.22 \\
% ours
  &  &   &   \\ %72.52 \\ 
% ours
  &  &   &   \\ %72.52 \\ 
  &  &   &   \\ %72.52 \\ 
  &  &   &   \\ %72.52 \\ 
  &  &   &   \\ %72.52 \\ 
  &  &   &   \\ %72.52 \\ 
\midrule
\multirow{16}{*}{Mobile-Former} & \multirow{16}{*}{\makecell{It refers to operators \\ within the Mobile-Former Block \\ that transition bi-directionally \\ between local and global representations}} & \multirow{16}{*}{\texttt{\makecell{features.1.local global.proj ->
features.1.global block.ffn.0 \\
features.1.global local.proj ->
features.2.conv1.0 \ \ \ \ \ \ \ \ \ \ \ \\
features.2.local global.proj -> 
features.2.global block.ffn.0 \\
features.2.global local.proj ->
features.3.conv1.0 \ \ \ \ \ \ \ \ \ \ \ \\
features.3.local global.proj -> 
features.3.global block.ffn.0  \\
features.3.global local.proj ->
features.4.conv1.0 \ \ \ \ \ \ \ \ \ \ \ \\
features.4.local global.proj -> 
features.4.global block.ffn.0 \\
features.4.global local.proj ->
features.5.conv1.0 \ \ \ \ \ \ \ \ \ \ \ \\
features.5.local global.proj ->
features.5.global block.ffn.0 \\
features.5.global local.proj ->
features.6.conv1.0 \ \ \ \ \ \ \ \ \ \ \ \\
features.6.local global.proj ->
features.6.global block.ffn.0 \\
features.6.global local.proj ->
features.7.conv1.0 \ \ \ \ \ \ \ \ \ \ \ \\
features.7.local global.proj -> 
features.7.global block.ffn.0 \\
features.7.global local.proj ->
features.8.conv1.0 \ \ \ \ \ \ \ \ \ \ \ \\
features.8.local global.proj -> 
features.8.global block.ffn.0}}} & \multirow{16}{*}{15} \\ %72.52 \\ 
  &  &   &   \\ %72.52 \\ 
  &  &   &   \\ %72.52 \\ 
  &  &   &   \\ %72.52 \\ 
  &  &   &   \\ %72.52 \\ 
  &  &   &   \\ %72.52 \\ 
  &  &   &   \\ %72.52 \\ 
 &  &   &   \\ %72.52 \\ 
 &  &   &   \\ %72.52 \\ 
 &  &   &   \\ %72.52 \\ 
 &  &   &   \\ %72.52 \\ 
 &  &   &   \\ %72.52 \\  
 &  &   &   \\ %72.52 \\  
 &  &   &   \\ %72.52 \\  
 &  &   &   \\ %72.52 \\  
 &  &   &   \\ %72.52 \\  
 
\midrule
\multirow{3}{*}{EfficientFormerV1} & \multirow{3}{*}{\makecell{These are operators that convert feature maps \\ 
from 4D Meta Block to 3D Meta Block}} & \multirow{3}{*}{\texttt{\makecell{stages.2.blocks.5.mlp.fc2 ->
stages.3.downsample.conv -> \\
stages.3.blocks.0.mlp.fc1}}} & \multirow{3}{*}{1} \\ %72.52 \\
  &  &   &   \\ %72.52 \\ 
  &  &   &   \\ %72.52 \\ 
\midrule
\multirow{4}{*}{EfficientFormerV2} & \multirow{4}{*}{\makecell{The bridge block consists \\ of local and global transition \\ operators exist in the 3rd and 4th stages}}  & \multirow{4}{*}{\texttt{\makecell{stages.2.downsample.conv.conv \ \ \ \ \ ->
stages.2.blocks.0.mlp.fc1.conv \\
stages.3.downsample.attn.proj.conv ->
stages.3.blocks.0.mlp.fc1.conv}}} & \multirow{4}{*}{2} \\ %72.52 \\ 
  &   &   &   \\ %72.52 \\ 
  &   &   &   \\ %72.52 \\ 
  &   &   &   \\ %72.52 \\ 
\bottomrule
\end{tabular}
}
\end{table*}

%\section{Bridge Block Details}

\secondR{
The term \texttt{bridge block} refers to the transitional part that connects both convolution and transformer blocks. This block's precise configuration changes significantly among hybrid ViT structures. Table~\ref{table:bridge_block} lists detailed descriptions and operator names for each model.
For MobileFormer, EfficientFormerV1, and EfficientFormerV2, the exact names of the operators that compose the bridge block vary slightly depending on the size of the model. The names shown in Table~\ref{table:bridge_block} are based on the smallest model size.
}

%Inductive bias (CNN+Transformer)
%MobileViTv1
%MobileViTv2: linear attention to reduce the complexity$O(N^2)$

\subsection{Hybrid Vision Transformer Quantization}
Uniform quantization is a commonly used method to quantize neural networks, including convolution networks and transformers.
As shown in Figure~\ref{fig:overview}, quantization for hybrid vision transformers is divided into three parts: convolution, bridge block, and transformer.
In uniform quantization, the weights and input activations are evenly quantized by each scale factor as: 
\begin{equation}
\vb{x} _{q} = \mathcal{Q}(\vb{x} _{r} ) = \text{clip}(\text{round} \left(\frac{\vb{x}_{r}}{\Delta_{\vb{x}}}+zp \right), \text{min}, \text{max}),
\label{lab:quantizer}
\end{equation}
where $\vb{x}_r$ is a real value (full precision) and $\vb{x}_q$ is a quantized value. $\Delta_{\vb{x}}$ is a scaling factor that is calculated depending on the quantization scheme: either asymmetric or symmetric. 
Also, $zp$ denotes the zero point and exists only when using the asymmetric scheme.

In the case of transformers, input data is first passed through a quantized embedding layer before entering the transformer blocks, which consist of a multi-head self-attention (MHSA) and a feed-forward network (FFN). 
The MHSA module computes queries $\vb{Q}$, keys $\vb{K}$, and values $\vb{V}$ with their pre-trained weights $\vb{W}$ and inputs $\vb{X}$.

In a given quantized multi-head self-attention layer (MHSA), the embedding matrix $\vb{E}$ undergoes quantization to $\vb{\bar{E}}$ prior to being processed by linear projection layers. These projection layers are represented as $\vb{\bar{Q}} = \vb{\bar{w}}_{Q}\vb{\bar{E}}$, $\vb{\bar{K}} = \vb{\bar{w}}_{K}\vb{\bar{E}}$, and $\vb{\bar{V}} = \vb{\bar{w}}_{V}\vb{\bar{E}}$, where the quantized weights $\vb{\bar{w}_{Q}}$, $\vb{\bar{w}}_{K}$, and $\vb{\bar{w}}_{V}$ correspond to the $\vb{Q}$ (Query), $\vb{K}$ (Key), and $\vb{V}$ (Value) projection layers, respectively.

\minorsecondR{
The subsequent step in the multi-head self-attention operation involves the use of divided query, key, and value matrices: $\mathbf{\bar{Q}}_h = \mathbf{\bar{Q}}/h$, $\mathbf{\bar{K}}_h = \mathbf{\bar{K}}/h$, and $\mathbf{\bar{V}}_h = \mathbf{\bar{V}}/h$. Each of these matrices is divided by the number of heads ($h$) to facilitate the computation for each individual head.
Hence, the multi-head self-attention operation is represented as follows:
\begin{align}
\text{MHSA}(\vb{\bar{Q}},\vb{\bar{K}},\vb{\bar{V}}) &= \text{concat}(head_i, head_{i+1},...,head_n)\vb{\bar{W}}^{O} \nonumber \\
head_i &= \text{quant-attention}(\mathbf{\bar{Q}}_h, \mathbf{\bar{K}}_h, \mathbf{\bar{V}}_h)
\label{eq:multi_atten}
\end{align}
}
\minorsecondR{
In Eq.~\eqref{eq:multi_atten}, the MHSA operation consolidates the results from each self-attention computation through concatenation followed by a linear projection to produce the final output. The quantized weight matrix used for the linear projection of the concatenated output is $\vb{\bar{W}}^{O}$. 
By combining multiple heads, the multi-head self-attention enables the attention function to extract information from different representation sub-spaces, which would be unattainable with a single attention head.}

%Following this, the multi-head self-attention mechanism utilizes $\vb{Q}$, $\vb{K}$, $\vb{V}$ along with softmax normalization to compute attention. 

\minorsecondR{
The detailed calculation of \textit{quant-attention} is described as 
\begin{align}
\text{quant-attention}&(\mathbf{\bar{Q}}_h, \mathbf{\bar{K}}_h, \mathbf{\bar{V}}_h)  \nonumber \\ 
&= \text{quant-softmax} \left( \frac{\vb{\bar{Q}}_h \times \vb{\bar{K}}^{T}_h}{\sqrt{\vb{d_{k}}}} \right)\times\vb{\bar{V}}_h,
\label{eq:quant_atten}
\end{align}
where $d_k$ denotes the key vector dimension.
In Eq.~\eqref{eq:quant_atten}, the scaling factor ($\sqrt{\vb{d_{k}}}$) is used to mitigate the issue of dot products becoming excessively large when $d_k$ is large. This size increase can cause the Softmax function to produce very small gradients, which in turn could result in the well-known vanishing gradient issue. Thus, the scaling factor effectively reduces the magnitude of the dot product outcomes, averting this complication.
After the MHSA layer, FFN takes a quantized output, which is concatenated from the results of MHSA as input.}

A popular method to reduce quantization error in post-training is reconstruction error minimization~\cite{yuan2022ptq4vit,adaround20,brecq21,apq22}.
Previous works have focused on optimizing the task loss, $\mathcal{L}=\text{Cross  Entropy}(\hat{\vb{y}},\vb{y})$, where $\hat{\vb{y}}$ represents the quantized output and $\vb{y}$ denotes the full precision output which is used as ground truth in PTQ. 
The expectation of the task loss is a function of network parameters $\vb{w}$, given by $\E[\mathcal{L}(\vb{x},\vb{y},\vb{w})]$, where $\vb{x}$ denotes activation and $\vb{y}$ denotes output. 
Quantization introduces a small perturbation $\epsilon$ on the parameter $\hat{\vb{w}} = \vb{w} + \epsilon$. 
Following the prior works~\cite{yuan2022ptq4vit,adaround20,brecq21,apq22}, we calculate the influence of quantization on the task loss using Taylor series expansion as:
\begin{equation}
    \E[\mathcal{L}(\vb{x},\vb{y},\hat{\vb{w}})] - \E[\mathcal{L}(\vb{x},\vb{y},\vb{w})] \approx \epsilon^{\intercal} \bar{g}^{(\vb{w})} + \frac{1}{2}\epsilon^{\intercal} \bar{H}^{(\vb{w})}\epsilon.
    \label{eq:taylor}
\end{equation}

Since the weight perturbation $\epsilon$ is relatively small, a second-order term of Taylor expansion can be used. 
In this equation, $\bar{g}^{(\vb{w})} = \E[\nabla_{\vb{w}}\mathcal{L}(\vb{x},\vb{y},\hat{\vb{w}})]$ is the gradient and can be ignored if the pre-trained model is well-converged. 
$\bar{H}^{(\vb{w})} = \E[\nabla_{\vb{w}}^2\mathcal{L}(\vb{x},\vb{y},\hat{\vb{w}})]$ is the Hessian matrix. 
The goal is to find a quantizer that includes optimal scaling factors or a rounding scheme to minimize the influence, given by $\min{\E[\mathcal{L}(\vb{x},\vb{y},\hat{\vb{w}})] - \E[\mathcal{L}(\vb{x},\vb{y},\vb{w})]}$. 
However, directly minimizing the task loss leads to overfitting problems due to small datasets during the calibration phase. 
Thus, the second-order term of the Taylor series (Eq.\eqref{eq:taylor}) is used. 
Referring to BRECQ~\cite{brecq21}, to reduce computational cost, Eq.(\eqref{eq:taylor}) is simplified by removing the gradient ($\bar{g}^{(\vb{w})}$) and approximating $\epsilon=\Delta \vb{w}$ to the network output ($\Delta O = \hat{O}-O$) as: \looseness=-1
\begin{equation}
    \epsilon^{\intercal} \bar{H}^{(x)}\epsilon \approx \Delta O^{\intercal}\bar{H}^{(O)}\Delta O.
    \label{eq:rem_basic}
\end{equation}
Referring to previous works~\cite{yuan2022ptq4vit,adaround20,brecq21,apq22}, MSE minimization based on the squared gradient that approximates the Hessian matrix captures the trend of the task loss more accurately than other metrics such as MSE, Cosine, and Pearson.

We adopt the methodology described in~\cite{yuan2022ptq4vit,apq22,wu2020easyquant} to traverse a search space of scaling factors by linearly dividing the maximum-minimum range of $\vb{w}$ and $\vb{x}$ into $n$ candidates as:
\begin{equation}
\begin{aligned}
    &[\alpha \frac{MAX|\vb{w}_{l}|}{2^{k-1}}, \beta \frac{MAX|\vb{w}_{l}|}{2^{k-1}}] \\ 
    &[\alpha \frac{MAX|\vb{x}_{l}|}{2^{k-1}}, \beta \frac{MAX|\vb{x}_{l}|}{2^{k-1}}],
\end{aligned}
\label{eq:candiates}
\end{equation}
where $\alpha$ and $\beta$ are utilized to control the number of candidates generated for scaling factors.

\iffalse
\begin{figure}[t]
\centering
%\subfloat[][]{\label{sub:minmax_layer}\includegraphics[width=.31\textwidth]{./figures/minmax_layer_quant}} %\vfill
\subfloat[][]{\label{sub:box_layer}\includegraphics[width=.4\textwidth]{./figures/15_channel_boxplot}} \vfill
\subfloat[][]{\label{sub:hist_layer}\includegraphics[width=.4\textwidth]{./figures/17_channel_boxplot}}
\caption{Per activation channel ranges of (a) the point-wise (\texttt{stage0.conv1x1.point.act}) and (b) depth-wise convolution (\texttt{stage0.convkxk.depth.act}) in the inverted bottleneck block in MobileViTv1-xxs}
\label{fig:layer_quant}
\end{figure}

\begin{figure}[t]
\centering
%\subfloat[][]{\label{sub:minmax_channel}\includegraphics[width=.33\textwidth]{./figures/minmax_channel_quant}} %\vfill
\subfloat[][]{\label{sub:hist_channel1_quant}\includegraphics[width=.4\textwidth]{./figures/17_hist_ch1}} \vfill
\subfloat[][]{\label{sub:hist_channel5_quant}\includegraphics[width=.4\textwidth]{./figures/17_hist_ch5}}
\caption{An overlapping histogram of quantized values (blue) and real values (orange) in the two channels of depth-wise convolution (\texttt{stage0.convkxk.depth.act}) in MobileViTv1-xxs: (a) the first channel and (b) fifth channel}
\label{fig:channel_quant}
\end{figure}
\fi

\subsection{\secondR{Challenges of Hybrid ViT Quantization}}
\label{sec:chall}
% Describe lessons 
%% Scheme: asymmetric problem
%% granularity: activation channel wise 

Here, we identify four critical challenges (C1--C4) that impede the quantization of hybrid ViTs and also explain why the current quantization method is insufficient.

\subsubsection{C1: Highly Dynamic Activation Range}
\label{subsec:C1}
To keep the accuracy from dropping too much when activation ranges change much, the quantization granularity should be automatically changed based on how the channels are spread out in each layer.
For layers that exhibit different ranges per channel, a scaling factor per channel could be chosen to preserve a specific layer~\cite{wu2020integer}.
Otherwise, channels exhibiting a narrow range might encounter the problem where all values are treated as zeros during the application of layer-wise quantization.

\begin{figure}[t]
	\centering
	\includegraphics[width=1\columnwidth]{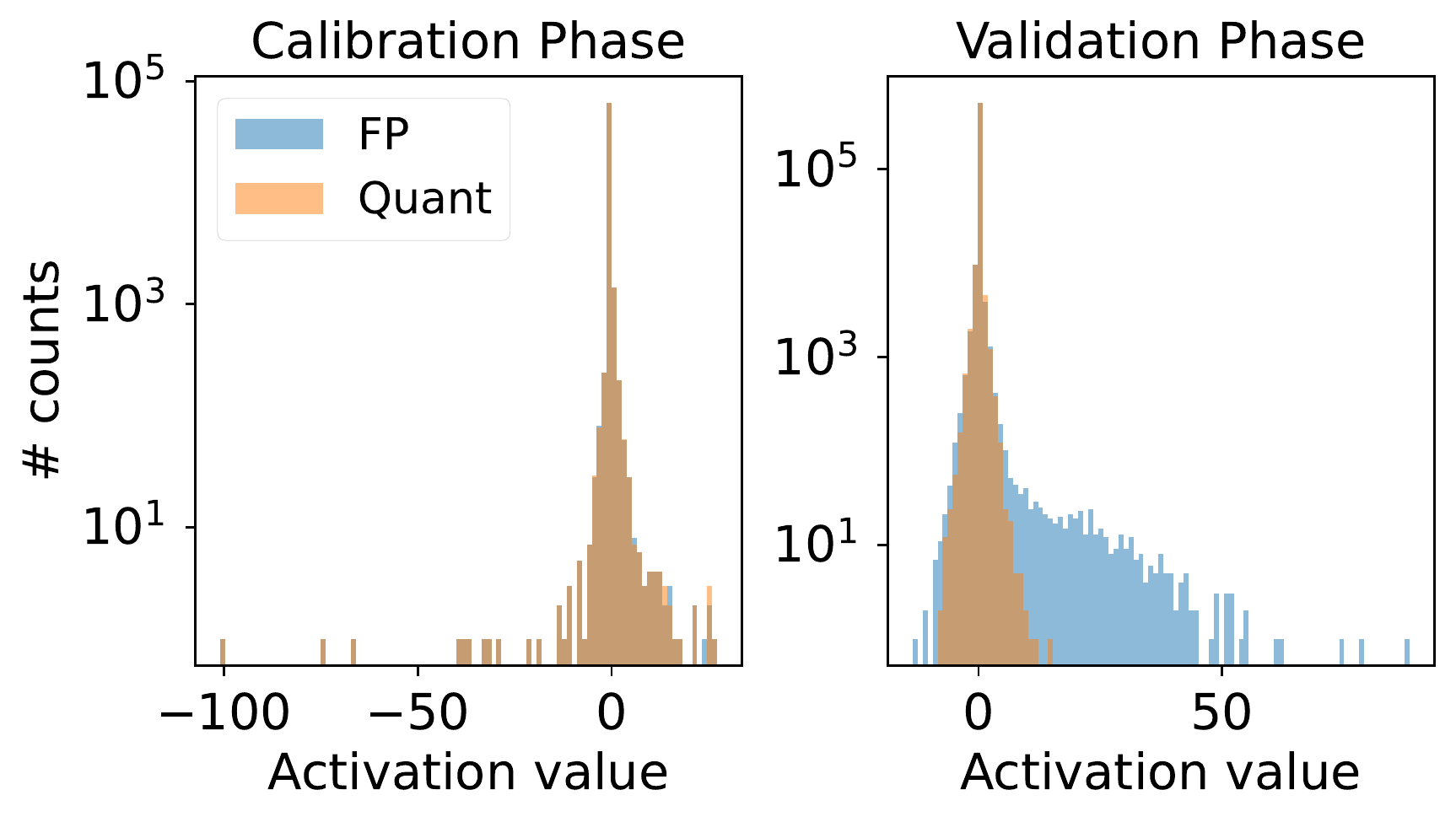}
	\caption{Discrepancy in activation ranges between the calibration and validation datasets in 1st bridge block of MobileViTv2-100}
 %Problematic activation channel (index 16) of convolution in bridge block due to overflow of zero point when using channel-wise manner and asymmetric scheme} 
	\label{fig:gap}
\end{figure}

To this end, channel-wise granularity is good for highly dynamic activations across channels.
When prioritizing accuracy preservation without considering latency, opting for a fine-grained granularity at the channel level could yield the highest accuracy.
However, \emph{this does not always hold true in the hybrid vision transformers.} 
%Also, fine-grained granularity (channel-wise) is to produce better quantization results than coarse-grained manner (layer-wise). 
Applying channel-wise quantization to every layer rather causes the scaling factors to overfit to small calibration data. This exacerbates the disparity between validation and calibration, resulting in a severe accuracy drop.
Figure~\ref{fig:gap} shows the phenomenon in which scaling factors, determined channel-wise during the calibration phase, exhibit discrepancies during the validation phase.
%Second, in the calibration level, a zero distribution channel appears.
As a result, simply applying channel-wise granularity for quantization across all layers is problematic in hybrid ViTs. 
In layers where overfitting poses a concern, applying the scaling factor on a layer-wise basis can alleviate this issue. Therefore, determining the optimal granularity of the scaling factor is a critical consideration.

\begin{figure}[t]
	\centering
	\includegraphics[width=1\columnwidth]{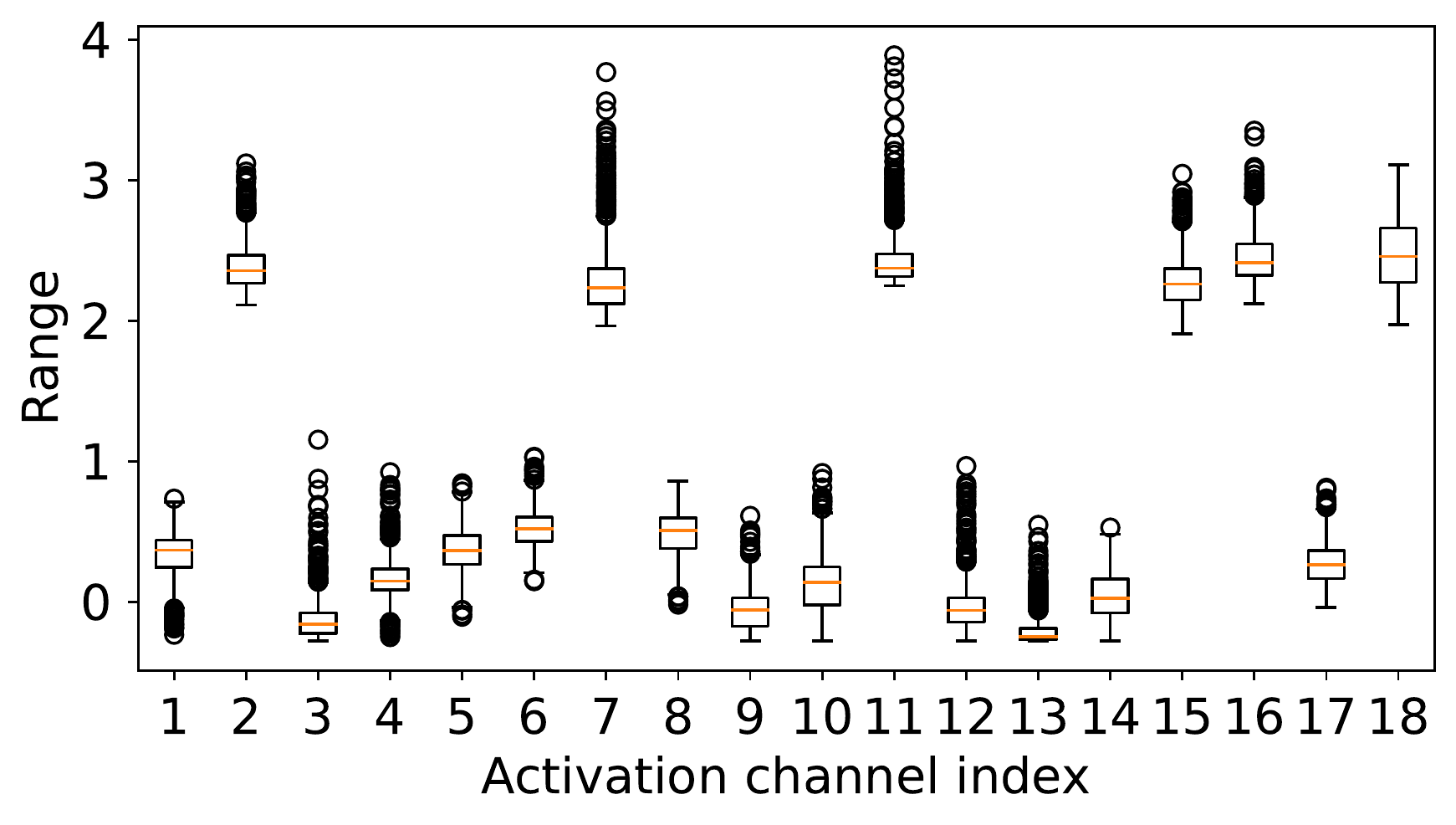}

	\caption{Per activation channel ranges of convolution in bridge block of MobileViTv1-xxs} 
 %make all values of channel index 1 to zero.} 
	\label{fig:small_range_clip}
\end{figure}

\subsubsection{C2: Zero-point Overflow in Bridge Block}
\label{subsec:C2}
Uniform quantization comprises two types: asymmetric and symmetric quantization, each with its unique pros and cons. 
In many cases, asymmetric quantization shows better accuracy than symmetric quantization. 
However, we find out that \emph{a severe accuracy drop occurs in the bridge block when using the asymmetric scheme or channel-wise granularity} due to highly dynamic activation ranges according to each channel and non-zero distribution. 

This observation aligns with the prior discovery that the {adoption of fine-grained granularity, such as channel-wise, does not always lead to a minimal quantization error.}
%specifically because the distribution is positively skewed, leading to an overflow. 

As shown in Figure~\ref{fig:small_range_clip}, the activation of the bridge block convolution shows a similar range for both the maximum and minimum values across all channels. This indicates that layer-wise quantization does not lead to significant accuracy degradation. 
However, when applying channel-wise quantization to distributions such as the 2nd, 7th, 11th, and so on in Figure~\ref{fig:small_range_clip}, where all values are greater than zero, overflow of the zero-point value for asymmetric quantization occurs (i.e., the zero-point value exceeded between $-128$ and $127$). 
As shown in Figure~\ref{fig:scheme_overflow}, the clipped zero point is used, resulting in certain values being reconstructed as a single value.

\begin{figure}[t]
	\centering
	\includegraphics[width=1\columnwidth]{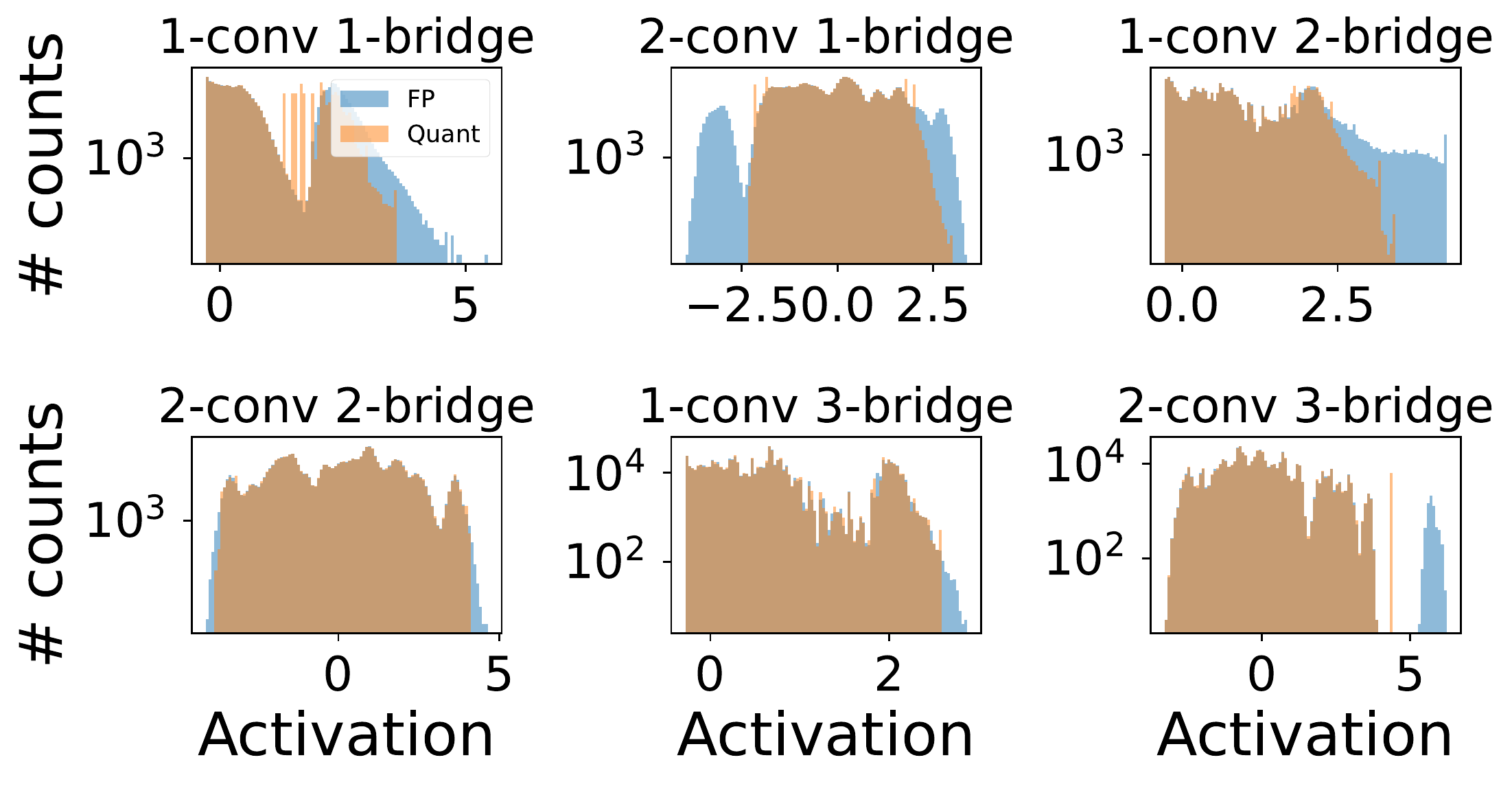}
	\caption{A histogram depicting the overlap between quantized values (blue) and real values (orange) for six activation layers in the 1st, 2nd, and 3rd bridge blocks of the MobileViTv1-xxs model} 
	\label{fig:xxs_overflow_granularity}
\end{figure}

\begin{figure}[t]
	\centering
	\includegraphics[width=1\columnwidth]{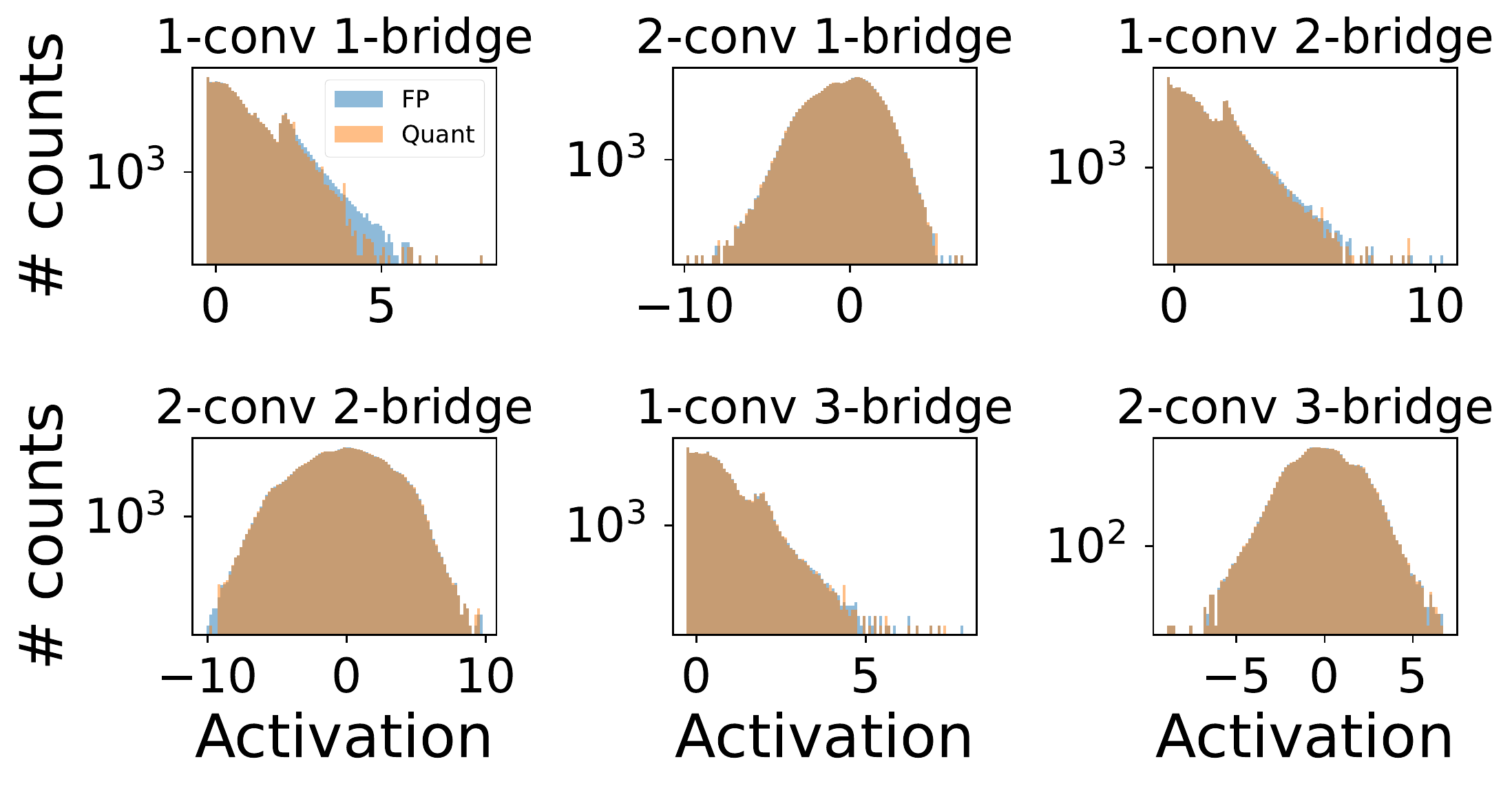}
	\caption{A histogram depicting the overlap between quantized values (blue) and real values (orange) for six activation layers in the 1st, 2nd, and 3rd bridge blocks of the MobileViTv1-xs model} 
	\label{fig:xs_overflow_granularity}
\end{figure}

\begin{figure}[t]
	\centering
	\includegraphics[width=1\columnwidth]{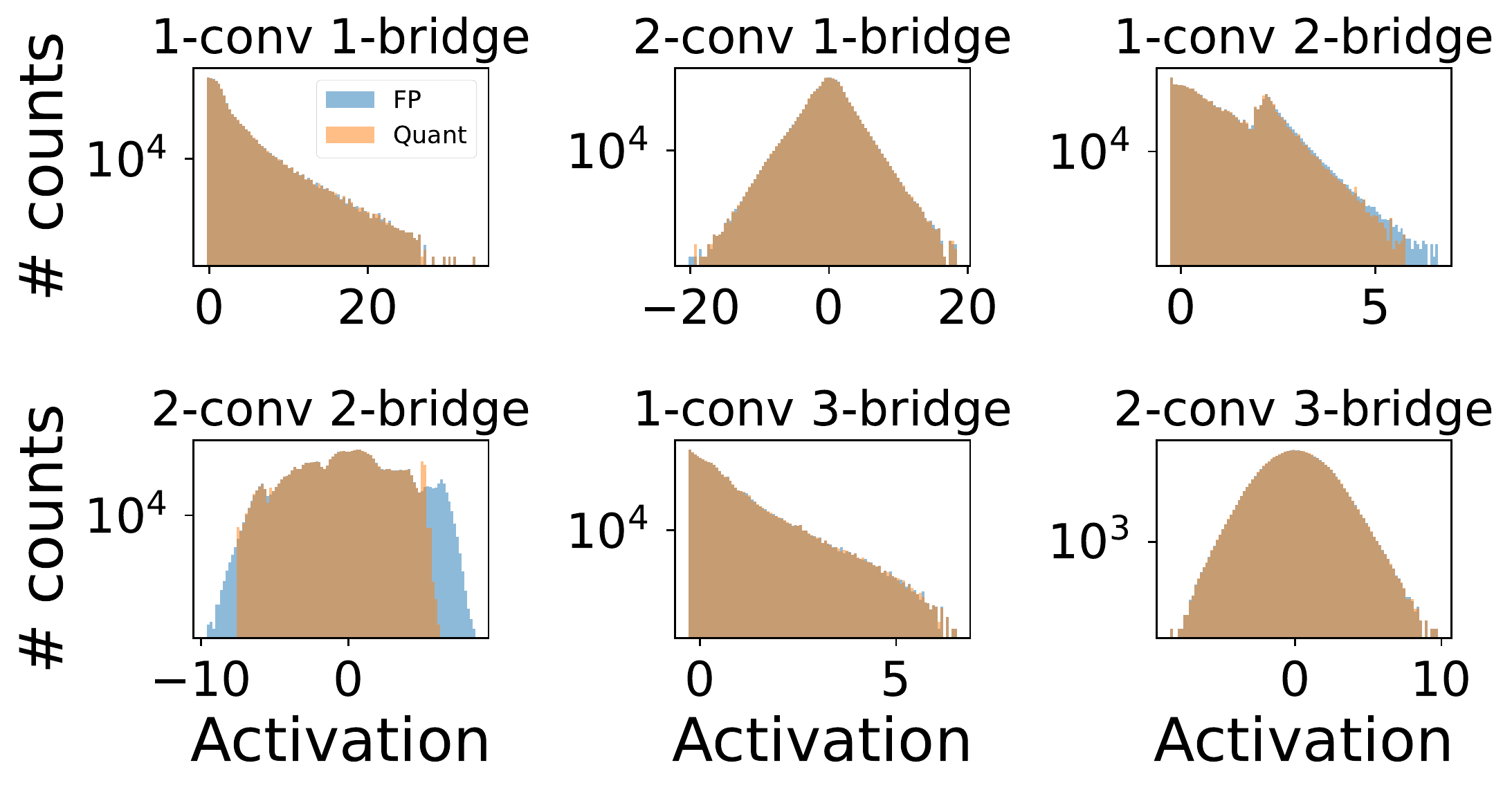}
	\caption{A histogram depicting the overlap between quantized values (blue) and real values (orange) for six activation layers in the 1st, 2nd, and 3rd bridge blocks of the MobileViTv1-s model} 
	\label{fig:s_overflow_granularity}
\end{figure}

\secondR{As shown in Figures~\ref{fig:xxs_overflow_granularity}, \ref{fig:xs_overflow_granularity}, and \ref{fig:s_overflow_granularity}, the issue of zero point overflow in activations within these bridge blocks is a phenomenon that exists across all bridge blocks in the MobileViT series (xxs, xs, and x).
In addition, we observe that the impact of zero point overflow diminishes as our models increase in size. 
This is because the influence of zero point overflow decreases as the number of channels expands, as in the cases of 64 (xxs), 96 (xs), and 144 (s).
Regardless of the model size, it is clear that clamping of specific values continues to occur in the bridge block. 
These issues manifest differently across models, necessitating an automated approach to choosing granularity and scheme.
}

\begin{figure}[t]
	\centering
	\includegraphics[width=1\columnwidth]{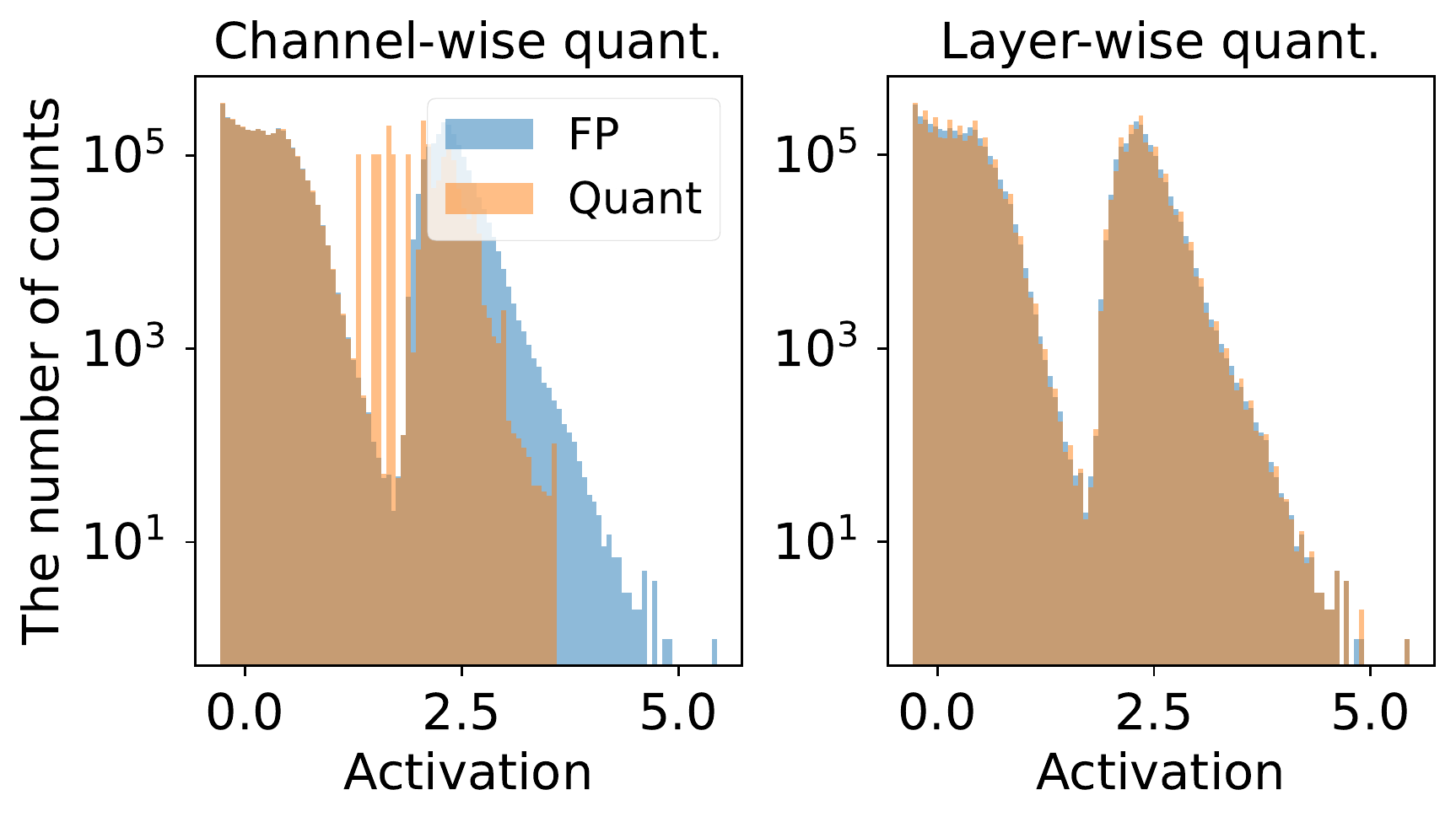}
	\caption{An overlapping histogram of quantized values (blue) and real values (orange) in the activation of  for 1st bridge block of MobileViTv1-xxs: (left) channel-wise quantization (right) layer-wise quantization} 
	\label{fig:overflow_granularity}
\end{figure}

\secondR{
Furthermore, for smaller models, we observe that the clamping issue associated with zero point overflow can be alleviated by transitioning from channel-wise quantization to layer-wise quantization, as demonstrated in Fig.~\ref{fig:overflow_granularity}.
In the end, the reason why the zero point overflows are clamped is due to the following:
\begin{align}
-128 &> q_{min} -\frac{r_{min}}{s} \nonumber \\
0 &> -\frac{r_{min}}{s} \nonumber \\ 
0 &< \frac{r_{min}}{s}
\label{eq:zero_proof}
\end{align}
In Eq.~\eqref{eq:zero_proof}, $q_{min}$ represents the minimum value of 8-bit quantization, which is $-128$, $s$ is the scaling factor, and $r_{min}$ refers to the minimum value of the original activation. As shown in Eq.~\eqref{eq:zero_proof}, since $s$ is always a positive value, if $r_{min}$ is greater than $0$, the zero point exceeds the range of $q_{min}$ and becomes clamped.
Therefore, when quantizing on a per-channel basis, if activations are composed only of values greater than or equal to $0$, it significantly causes a decrease in accuracy. 
To solve this issue, using layer-wise quantization takes into account the entire layer, including values less than or equal to $0$ as $r_{min}$. 
Consequently, the zero-point value falls within the 8-bit range ($-128$ to $127$).
}

%In this section, we demonstrate that the zero point overflow issue arises across multiple bridge blocks within each hybrid vision transformer model.
%Two problematic activations are present in every bridge block.
%Both MobileViTv1 and MobileViTv2 comprise three bridge blocks.
%The distribution of activations before and after quantization for the convolution operations in each bridge block is illustrated in Fig.\ref{fig:xxs_overflow_granularity}, Fig.\ref{fig:xs_overflow_granularity}, and Fig.~\ref{fig:s_overflow_granularity}.

\iffalse
\begin{figure}[t]
	\centering
	\includegraphics[width=.7\columnwidth]{./figures/59_hist_ch16}
	\caption{Problematic activation channel (index 16) of convolution in bridge block due to overflow of zero point when using channel-wise manner and asymmetric scheme} 
	\label{fig:scheme_overflow}
\end{figure}
\fi 
\begin{figure}[t]
	\centering
	\includegraphics[width=1\columnwidth]{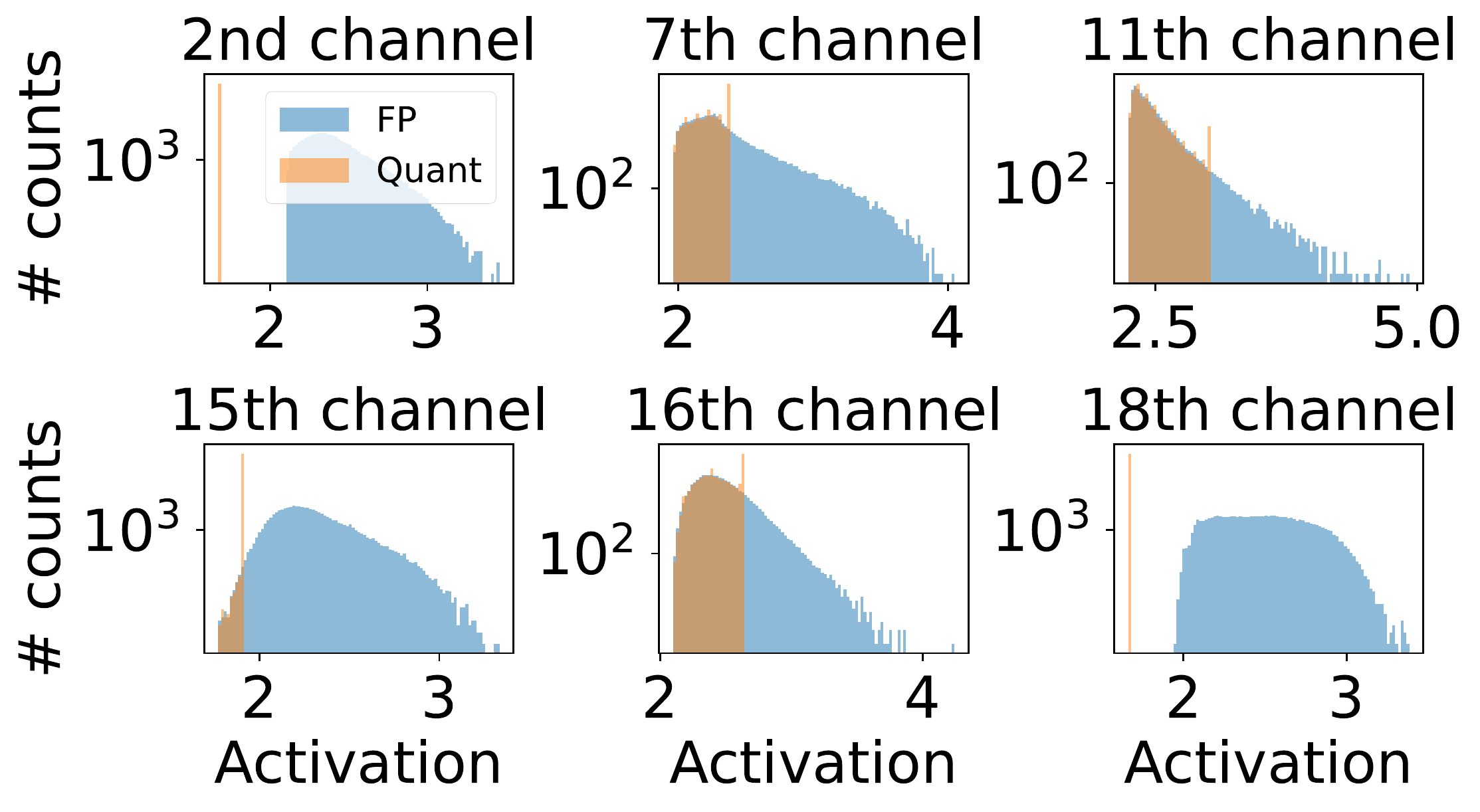}
	\caption{The selected problematic activation channels of convolution in bridge block of MobileViTv1-xxs due to overflow of zero point when using the channel-wise manner and asymmetric scheme} 
	\label{fig:scheme_overflow}
\end{figure}

\subsubsection{C3: Quantization with Diverse Normalizations} \label{subsec:C3}%% Batch-Norm, I-Norm, G-Norm 
\emph{Hybrid vision transformers, unlike pure models such as CNN and ViT, employ different combinations of normalization techniques.} 
In the case of MobileViTv1, Mobile-Former, EfficientFormerV1, and EfficientFormerV2, {BatchNorm} and {LayerNorm} are utilized, while MobileViTv2 uses {BatchNorm} and {GroupNorm}.
To build a network as parameter-efficient as possible, various normalization techniques are employed in each hybrid vision transformer. However, a method to quantize all the normalization techniques used has yet to be proposed. 
In detail, unlike {BatchNorm}, the computation of {LayerNorm}, {GroupNorm}, and {InstanceNorm} requires dynamic computation to determine the mean and variance. 
When dynamic operations are handled in a floating point, additional data movement occurs in off-chip memory. 
Therefore, to minimize the inference latency, the normalization should be quantized so that as many dynamic operations as possible can be processed in the integer domain.
%To address this, prior work~\cite{lin2022fq} introduced {I-layerNorm} for integer domain computation.
%In this work, we propose \textit{I-GroupNorm}, which is specifically designed for fully quantized MobileViTv2 and can also be computed under the integer domain.
% how to extend I-layerNorm to I-GroupNorm 

\subsubsection{C4: Sub-5M Parameter Models} \label{subsec:C4}%% analysis depending on model size
Compared to heavy CNN models such as ResNet and VGG series, lightweight models with a relatively small number of parameters and few residual connections are more susceptible to the quantization process. \emph{This vulnerability is particularly pronounced in models that have fewer than 5M parameters, where the quantization error is significantly higher in hybrid vision transformers.} As previously mentioned, using an asymmetric scheme for an activation distribution with a minimum value of $0$ or greater can lead to force clamping due to the zero point overflow problem, making the accuracy more sensitive to the quantization granularity and scheme.

%구체적으로, zero point overflow를 발생 시키는 activation은 MobileViT 모델 계열의모든 brdige block에 2개씩 존재하며, 모델 사이즈에 따라 정확도 하락의 폭이 크게 다르다.

\section{Methodology}
Addressing the four identified challenges (C1--C4), we design Q-HyViT, a precise post-training quantization framework specifically tailored for hybrid ViTs. It automatically identifies optimal layer-wise strategies -- \news{choosing granularity (channel-wise or layer-wise) for C1, C2, and C4, quantization scheme (symmetric or asymmetric) for C2 and C4,} and quantization scaling factor for C3 and C4. This is achieved by leveraging the proposed hybrid reconstruction error minimization method.

%the strategic selection of optimal scale factors, granularity, and scheme for both the bridge block and individual layers based on reconstruction error from a loss degradation perspective.

%As shown in Figure~\ref{fig:overview}, we describe Q-HyViT that minimizes the quantization error of hybrid vision transformers by selecting optimal scale factors, granularity, and scheme for both the bridge block and each layer based on reconstruction error from a loss degradation perspective.

% hessian guided search in convolution
% -- loss 분석, quantized loss (sequential gradient analysis) 

\subsection{Hybrid Reconstruction Error Minimization}
Previous post-training quantization methods for ViTs have typically aimed to optimize the quantization task loss, which is achieved through the utilization of reconstruction error minimization, relying on second-order metrics to assess potential scaling factors for quantization. Furthermore, these methods incorporate a weight-rounding mechanism to enhance the overall quantization process. 
However, faced with the challenges specific to hybrid vision transformers, they become less effective in dealing with \emph{bridge blocks}, which contain a gap between local and global representation, and \emph{high dynamic activation ranges} resulting from the mixed structure of CNN and transformer.

% ------- It is better to insert figures to show evidence. like loss plane.
%In an ideal scenario, a carefully designed loss term would determine how to quantize each layer with the lowest quantization loss to converge to a global optimum. 
%Nonetheless, we observed that the second-order loss terms calculated by each candidate exhibit significant variance, especially at the bridge layer.
%Consequently, we propose bridge-wise reconstruction error minimization as a post-training quantization technique. This technique optimizes calibration on a bridge-wise basis, enabling the Hessian-guided loss to perceive the quantization error of neighboring layers in a transition block from CNN to a transformer.

\news{
To address these issues, Q-HyViT introduces \emph{hybrid reconstruction error minimization}, the schematic of which is illustrated in Figure~\ref{fig:overview}. As depicted in Figure~\ref{fig:overview}, the Hybrid ViT architecture is divided into three distinct blocks for quantization: local, global, and bridge. The local blocks primarily consist of several convolutional layers, with the output from these layers denoted as $\hat{O}^l$. To mitigate overfitting issues on the small calibration dataset and alleviate high computational demands, the reconstruction error for each layer's output is calculated.
}

\news{
The global block, representing the transformer, computes the reconstruction error in a manner similar to the local block, using $\hat{O}^l$ as a reference. The bridge block serves as a transitional operator facilitating the integration between local and global processing. Specific operations within this block may vary slightly across different models. To alleviate issues related to the highly dynamic activation range within the bridge block, the reconstruction error is calculated based on the output of this block, denoted as $\hat{O}^{bb}$, taking into account all dependencies within the block to optimize the process.
}

\news{
The model’s quantization process encompasses both forward (illustrated by solid and dashed lines in Figure~\ref{fig:overview}) and backward (illustrated by dashed red lines in Figure~\ref{fig:overview}) passes. The diagram outlines the computation of a loss function $\mathcal{L}$, which is likely designed to measure the discrepancy between the full-precision output $y_{FP32}$ and the quantized output $\hat{y}_{w_{q}a_{q}}$. This loss function is utilized to update the model parameters during calibration to minimize quantization errors.
}

The proposed method distinguishes the reconstruction strategy based on whether a given layer is part of the bridge block or not. It then determines the appropriate granularity and quantization scheme for each layer during post-training quantization. The reconstruction objective ($O^{bb}$) of the hybrid approach can be represented as:
\begin{align}
O^{bb} &=
\begin{cases}
    \vb{w}_{n}^{bb}\vb{w}_{n-1}^{bb} \dots \vb{w}_{1}^{bb}\vb{x}^{bb}, & \text{if a layer is in a bridge block}\\
    \vb{w}^{\ell}\vb{x}^{\ell}, & \text{otherwise } bb \text{ is equal to } \ell   
\label{eq:layer_type} 
\end{cases} 
\end{align}
\firstR{In Eq.~\eqref{eq:layer_type}, $O^{bb}$ is determined to be either a single layer or multiple layers based on the presence of a bridge block.}
When a layer is part of the bridge block, the objective encompasses all the preceding layers within that bridge block.

Moreover, our hybrid reconstruction not only enables hybrid ViTs to achieve minimal quantization errors, but also determines quantization granularity and scheme automatically for each layer, using the hybrid reconstruction equation, guided by the reconstruction objective:
\begin{equation}
\begin{aligned}
    & \min\limits_{\Delta,g,s}{\E \left[\Delta O^{(bb),\intercal},\textbf{H}^{O^{(bb)}} \Delta O^{(bb)} \right]} \approx \\  
    & \min\limits_{\Delta,g,s}{\E \left[\Delta O^{(bb),\intercal}, \textit{diag} \left((\frac{\partial \textit{L}}{\partial O_{1}^{(bb)}})^2 ,\cdots, (\frac{\partial \textit{L}}{\partial O_{|O^{bb}|}^{(bb)}})^2 \right) \Delta O^{(bb)} \right]},
    \label{eq:rem}
\end{aligned}
\end{equation}

\smallskip\smallskip\smallskip
\noindent where $bb$ is $\in [BridgeBlock, a\text{ }layer]$.
In Eq.~\eqref{eq:rem}, $\Delta O^{(bb)}$ is the difference between the quantization outputs before and after.
$O_{n}^{(bb)}$ indicates the $n$-th element of $O^{bb}$. 
The range of $n$ is from $1$ to $O_{|O^{bb}|}$.
As a result, we optimize the optimal scaling factor ($\Delta \!\in\! [1,100]$), granularity~($g\!\in\! [layer,channel]$), and scheme~($s\in [asymmetric, symmetric]$) through hybrid reconstruction error minimization.

\subsection{\news{Implementing Q-HyViT: From Calibration to Optimization}}
The proposed Q-HyViT method, as outlined in Algorithm~\ref{algo:qhyvit}, leverages hybrid reconstruction to determine optimal scaling factors, granularity, and scheme. Thus, it effectively diminishes quantization errors in hybrid ViTs.

\news{
The algorithm~\ref{algo:qhyvit}, similar to PTQ methods, can be divided into two primary stages: calibration and quantization optimization. These stages are organized around three outermost \textsf{while} loops.
}

\news{
Throughout the calibration stage, Q-HyViT computes the output and gradient of each bridge block and layer via forward and backward propagation. These operations correspond to the first and second \textsf{while} loops in Algorithm~\ref{algo:qhyvit} (\textsf{\textit{lines 1--11}}). 
Specifically, the first \textsf{while} loop executes the model without quantization to calculate and store $\vb{y}^{(fp32)}$ and the intermediate results $O^{bb}_{l}$ of the layers. The second \textsf{while} loop performs the default quantization layer (min-max scaling) or bridge blockwise, then computes the loss based on the difference from the outputs of the first loop to perform backpropagation, storing the outputs and gradients of each layer and bridge block.
}

\news{
The third \textsf{while} loop represents the optimization of quantization across all hybrid transformer layers by minimizing reconstruction error (\textsf{\textit{lines 12--18}}). As mentioned, in hybrid ViTs, the dynamic activation range varies widely, necessitating automated adjustment methods. From this perspective, this approach minimizes quantization errors to enhance the second-order metric and ultimately leads to more accurate models.
}

\news{
To provide a more intuitive understanding, we describe how Algorithm~\ref{algo:qhyvit} addresses each challenge we identified:
\textit{(1) a highly dynamic activation range} (refer to \S\ref{subsec:C1}) is addressed in the third loop of Algorithm~\ref{algo:qhyvit}, which identifies optimal granularity (either channel-wise or layer-wise). As mentioned previously, an indiscriminate choice of channel granularity can lead to overfitting with a small calibration dataset; hence, channel-wise granularity is selected only when necessary, considering the activation range.
\textit{(2) zero-point overflow in bridge blocks} (See to \S\ref{subsec:C2}) is also resolved by optimal granularity selection and further alleviated by optimally choosing the quantization scheme. Zero-point overflow, a problem of exceeding the min-max range, can be mitigated by adjusting the scheme to symmetric or tuning the min-max values through granularity. These optimal choices are automatically made to minimize the reconstruction error.
\textit{(3) diverse normalization} (refer to \S\ref{subsec:C3}) is directly addressed by the proposed implementation of hybrid reconstruction error minimization. The proposed algorithm is implemented not only for layer normalization, but also extends to group and batch normalization, optimizing scaling for each.
\textit{(4) Quantization of sub-5M parameter models} inevitably leads to accuracy loss due to the lack of redundant parameters (See to \S\ref{subsec:C4}). Therefore, all the proposed components outlined in Algorithm~\ref{algo:qhyvit} are employed to ensure that our quantization preserves accuracy as much as possible. However, the quantization of extremely small-sized models remains a difficult challenge, necessitating further exploration of various approaches.
}

\IncMargin{1em}
\setlength{\textfloatsep}{10pt}% Remove 
\begin{algorithm}[t]
	\DontPrintSemicolon % Some LaTeX compilers require you to use \dontprintsemicolon instead
	\Indm
	\KwIn{A hybrid vision transformer model and a few images for calibration;}
	\KwOut{Optimal scaling factors($\Delta^{\ast}$) scheme($s$) and granularity($g$);}
	\Indp
        \While{a layer$(\ell)$ is not the end of layer}{
        \tcc{full-precision outputs on each layer including final layer $(\vb{y}^{(fp32)}$)}
        $O_{\ell}^{bb} \gets$ forward propagation $(\vb{w}_{\ell}\vb{x}_{\ell})$\;
	}
         \While{a layer$(\ell)$ is not the end of layer}{
            \If {a layer$(\ell)$ in a Bridge Block}{
              Backward propagation to get $\frac{\partial \mathcal{L}}{\partial O^{bb}}$\;
           }
            \Else{
                 Backward propagation to get $\frac{\partial \mathcal{L}}{\partial O^{\ell}}$\;
            }
	}
        \While{a layer$(\ell)$ is not the end of layer}{
            \tcc{initialize scaling factors}
            $\Delta^{\ast}_{\vb{w}_{\ell}^{bb}},\Delta^{\ast}_{\vb{x}_{\ell}^{bb}} \gets \frac{MAX(|\vb{w}_{\ell}^{bb}|)}{2^k}, \frac{MAX(|\vb{x}_{\ell}^{bb}|)}{2^k}$  \; 
            
            \tcc{Generate candidates for scaling factors} $\Delta_{\vb{w}_{\ell}^{bb}},\Delta_{\vb{x}_{\ell}^{bb}} \gets$ Eq.~\eqref{eq:candiates}\;
            \While{Three iterations}{
                \tcc{Determine granularity, scheme, scaling factors}
                $g,s,\Delta_{\vb{w}_{\ell}^{bb}}^{\ast},\Delta_{\vb{x}_{\ell}^{bb}}^{\ast} \gets$ Eq.~\eqref{eq:rem}\;
            }
        }
        \Return{$\Delta^{\ast}$ , $g$ , $s$}
	\caption{The tuning process of Q-HyViT}
	\label{algo:qhyvit}
\end{algorithm}
\DecMargin{1em}

\begin{table*}[t]
\centering
\caption{A comparison of three post-training quantization methods for image classification on ImageNet-1K using five hybrid ViT architectures and bit-widths. In quantized models, softmax and layer-norm remain under floating-point. 
%We abbreviate MobileViT as "MV", Mobile-Former as "MF", and EfficientFormer as "EF"
}
\label{table:accuracy}
%\resizebox{\columnwidth}{!}{
%\resizebox{\textwidth}{!}{
%\resizebox{11cm}{!}{
%\begin{tabular}{@{\hskip0pt}l@{\hskip0pt}c@{\hskip0pt}c@{\hskip2pt}c@{\hskip2pt}c@{\hskip2pt}c@{\hskip2pt}c@{\hskip2pt}c@{\hskip2pt}c@{\hskip2pt}c@{\hskip2pt}
\begin{tabular}{lccccccccccc
} %~\cite{migacz20178,wu2020integer}
\toprule
\multirow{2}{*}{\textbf{Model}} &
\multirow{2}{*}{\textbf{\# Params.}} &
\multirow{2}{*}{\textbf{Type}} &
\multirow{2}{*}{\textbf{FP32}} & 
%\multicolumn{1}{c}{\textbf{Liu21~\cite{liu2021post}}} &
\multicolumn{2}{c}{\textbf{EasyQuant}~\cite{wu2020easyquant}} &
%\multirow{2}{*}{\textbf{${}^\ast$FQ-ViT~\cite{lin2022fq}}} & 
\multicolumn{2}{c}{\textbf{PTQ4ViT}~\cite{yuan2022ptq4vit}} &
\multicolumn{2}{c}{\textbf{\firstR{RepQ-ViT}}~\cite{li2023repq}} &
\multicolumn{2}{c}{\textbf{Ours}} \\
 & & & & W8A8 & W6A6 & W8A8 & W6A6 & \firstR{W8A8} & \firstR{W6A6} & W8A8 & W6A6 
\\   \midrule
\midrule
MobileViTv1-xxs & 1.3M & Hybrid & 69.0    & 36.13 & 10.17 & 37.75 & 30.80  & \firstR{1.85} & \firstR{1.38}  & 68.20 & 66.33 \\ 
MobileViTv1-xs & 2.3M & Hybrid & 74.8     & 73.16  & 55.22 &  65.52 & 62.56  & \firstR{41.96} & \firstR{27.29}  & 74.31 & 73.44 \\ 
MobileViTv1-s & 5.6M & Hybrid & 78.4    & 74.21 & 42.70 &  68.19 & 65.07  & \firstR{59.01} & \firstR{56.61}  & 77.92  & 77.18 \\  
\midrule
MobileViTv2-050 & 1.4M & Hybrid & 70.2  & 66.80 & 11.58  & 39.39 & 45.38  & \firstR{26.60} & \firstR{27.89}  & 69.89 & 69.07 \\ %68.22 \\
% ours
MobileViTv2-075 & 2.8M & Hybrid & 75.6  & 62.91 & 2.54  & 65.54 & 65.85 & \firstR{55.52} & \firstR{40.61}  & 75.29 & 74.58 \\ %72.52 \\ 
% ours
MobileViTv2-100 & 4.9M & Hybrid & 78.1 & 69.34 & 0.12 & 51.02 & 47.27  & \firstR{40.85} & \firstR{26.07} & 77.63 & 77.11 \\ %72.52 \\ 
MobileViTv2-125 & 7.5M & Hybrid & 79.6 & 77.31 & 4.56 & 67.39 & 59.39  & \firstR{41.65} & \firstR{30.43} & 79.31 & 77.03 \\ %72.52 \\ 
MobileViTv2-150 & 10.6M & Hybrid & 80.4 & 75.83 & 10.39 & 68.61 & 67.58  & \firstR{62.12} & \firstR{58.71} & 80.09 & 79.97 \\ %72.52 \\ 
MobileViTv2-175 & 14.3M & Hybrid & 80.8 & 79.93 & 47.22 & 72.30 & 71.78  & \firstR{63.52} & \firstR{62.89} & 80.63 & 80.45 \\ %72.52 \\ 
MobileViTv2-200 & 18.5M & Hybrid & 81.2 & 80.04 & 57.32 & 75.50 & 74.65 & \firstR{64.65} & \firstR{62.15} & 80.94 & 80.76 \\ %72.52 \\ 
\midrule
Mobile-Former-26m & 3.2M & Hybrid & 64.0 & 28.95 & 0.12 & 58.27 & 47.25 & \firstR{0.11} & \firstR{0.16} & 61.78 & 51.06 \\ %72.52 \\ 
Mobile-Former-52m & 3.5M & Hybrid & 68.7 & 62.16 & 17.29 & 67.32 & 62.01 & \firstR{1.12} & \firstR{1.00}  & 67.79 & 62.65 \\ %72.52 \\ 
Mobile-Former-96m & 4.6M & Hybrid & 72.8 & 53.31 & 33.68 & 71.32 & 64.72 & \firstR{0.40} & \firstR{0.25}  & 71.60 & 64.21 \\ %72.52 \\ 
Mobile-Former-151m & 7.6M & Hybrid & 75.2 & 4.98 & 3.49 & 73.86 & 68.16 & \firstR{0.11} & \firstR{0.12}  & 74.30 & 68.44 \\ %72.52 \\ 
Mobile-Former-214m & 9.4M & Hybrid & 76.7 & 72.79 & 28.32 & 75.01 & 68.24 & \firstR{0.13} & \firstR{0.14}  & 75.76 & 69.34 \\ %72.52 \\ 
Mobile-Former-294m & 11.4 & Hybrid & 77.9 & 74.15 & 59.55 & 76.96 & 74.48 & \firstR{1.05} & \firstR{0.58}  & 76.93 & 74.6 \\ %72.52 \\ 
Mobile-Former-506m & 14.0M & Hybrid & 79.3 & 78.01 & 67.14 & 75.44 & 70.13 & \firstR{0.19} & \firstR{0.26}  & 75.60 & 74.67 \\ %72.52 \\ 
\midrule
EfficientFormerV1-L1 & 12.3M & MetaBlock & 80.2 & 78.24 & 58.83 & 80.11 & 79.8 & \firstR{80.36} & \firstR{78.55}  & 80.15 & 77.25 \\ %72.52 \\
EfficientFormerV1-L3 & 31.3M & MetaBlock & 82.4 & 82.39 & 80.38 & 82.39 & 82.36 & \firstR{82.41} & \firstR{82.29}  & 82.46 & 82.18 \\ %72.52 \\ 
EfficientFormerV1-L7 & 82.1M & MetaBlock & 83.3 & 83.24 & 81.89 & 83.34 & 83.16 & \firstR{83.28} & \firstR{83.03}  & 83.31 & 83.12 \\ %72.52 \\ 
\midrule
EfficientFormerV2-S0 & 3.5M & Hybrid & 76.2 & 68.21 & 41.24 & 68.40 & 41.26 & \firstR{40.02} & \firstR{37.11} & 74.69 & 74.18 \\ %72.52 \\ 
EfficientFormerV2-S1 & 6.1M & Hybrid & 79.7 & 66.42 & 2.69 & 73.44 & 73.34 & \firstR{58.30} & \firstR{53.06} & 77.56 & 77.54 \\ %72.52 \\ 
EfficientFormerV2-S2 & 12.6M & Hybrid & 82.0 & 71.80 & 7.02 & 79.85 & 79.39 & \firstR{70.39} & \firstR{70.37} & 80.62 & 80.30 \\ %72.52 \\ 
EfficientFormerV2-L & 26.1M & Hybrid & 83.5 & 80.34 & 3.34 & 82.46 & 82.22 & \firstR{76.72} & \firstR{74.33} & 82.80 & 82.71 \\ %72.52 \\ 
% ours
%MobileViTv2-100 & 4.3M & Hybrid & 78.09 & 69.34 & 0.12 & 61.74  & 0.1 \\ %0.1 \\ 
% ours
\bottomrule
\end{tabular}
%}
\end{table*}

\begin{table}[t]
\centering
\caption{Fully quantized accuracy of hybrid vision transformer architectures.}
\label{table:full_accuracy}
\resizebox{\columnwidth}{!}{
%\resizebox{\textwidth}{!}{
\begin{tabular}{@{\hskip0pt}l@{\hskip0pt}c@{\hskip0pt}c@{\hskip0pt}c@{\hskip0pt}c@{\hskip0pt}c@{\hskip0pt}c@{\hskip0pt}c@{\hskip0pt}c@{\hskip0pt}
} %~\cite{migacz20178,wu2020integer}
\toprule
\multicolumn{1}{c}{\textbf{Model}} &
\multicolumn{1}{c}{\textbf{\# Params.}} &
\multicolumn{1}{c}{\textbf{Type}} &
\multicolumn{1}{c}{\textbf{FP32}} & 
\multicolumn{1}{c}{\textbf{FQ-ViT}} & 
\multicolumn{1}{c}{\textbf{Ours}}
\\   \midrule
\midrule
MobileViTv1-xxs (MVv1-xxs) & 1.3M & Hybrid & 68.91 & 0.1 & 67.20 \\ 
% ours
MobileViTv1-xs (MVv1-xs) & 2.3M & Hybrid & 74.64 & 62.2 & 73.89 \\ 
% ours
MobileViTv1-s (MVv1-s) & 5.6M & Hybrid & 78.31 & 74.94 & 77.72 \\
% ours 
MobileViTv2-050 (MVv2-050) & 1.4M & Hybrid & 70.16 & 5.00 & 68.73 \\
% ours
MobileViTv2-075 (MVv2-075) & 2.8M & Hybrid & 75.62 & 34.60 & 74.36 \\
% ours
MobileViTv2-100 (MVv2-100) & 4.3M & Hybrid & 78.09 & 0.40 & 77.13 \\ 
% ours
\bottomrule
\end{tabular}
}
\end{table}

\section{\secondR{Experiments}}
We conduct extensive comparisons of Q-HyViT against various existing quantization methods. As discussed previously, there is no comprehensive method to quantize the hybrid vision transformers.
Therefore we directly implemented the following open-sourced state-of-the-art quantization algorithms for pure vision transformers, namely EasyQuant~\cite{wu2020easyquant}, FQ-ViT~\cite{lin2022fq}, PTQ4ViT~\cite{yuan2022ptq4vit}, \firstR{and RepQ-ViT~\cite{li2023repq}} and then applied them on five hybrid vision transformer architectures.

%In this section, we compared the proposed method with several quantization algorithms. 
%As we mentioned earlier, there is no comprehensive method to quantize the hybrid vision transformer.
%Therefore, we directly implemented the following open-sourced algorithms, EasyQuant~\cite{wu2020easyquant}, FQ-ViT~\cite{lin2022fq}, and PTQ4ViT~\cite{yuan2022ptq4vit}, on hybrid vision transformer.

\subsection{\secondR{Implementation Details}}
For a fair comparison, we maintained most configurations consistent with EasyQuant, PTQ4ViT, \minorsecondR{RepQ-ViT}, and FQ-ViT. Specifically, our settings vary depending on whether the model is fully quantized or not. Additionally, the five hybrid vision transformer models were referred to as the official models.

\subsubsection{\secondR{Model Download}}
\secondR{Except for Mobile-Former, the other four hybrid models leverage the \texttt{timm} framework~\footnote{https://github.com/huggingface/pytorch-image-models}. Meanwhile, Mobile-Former is built upon the implementation by AAboys~\footnote{https://github.com/AAboys/MobileFormer}. We successfully replicated the original accuracy using open-source codes under FP32 precision.}

\subsubsection{\secondR{Settings for EasyQuant and PTQ4ViT}}
\secondR{In EasyQuant, we quantized all operators, including fully-connected layers and matrix multiplications. To obtain the optimal scaling factors, we employed a search algorithm based on cosine distance. The search space was derived from $\alpha$ = 0.5 and $\beta$ = 1.2.}

\secondR{In PTQ4ViT, we adjusted the hyperparameters to $\alpha$ = 0 and $\beta$ = 1.2. Similar to PTQ4ViT, this study adopted the parallel quantization method to prevent a significant accuracy drop caused by small datasets.}

\secondR{In both cases, we selected a sample of 32 images from the training dataset during the calibration process.}

\subsubsection{\firstR{Settings for RepQ-ViT}}
\firstR{
To reproduce the results, we used the default settings from the code published by RepQ-ViT. During calibration, RepQ-ViT uses the percentile~\cite{li2019fully}. 
For quantization, we applied an asymmetric method with channel-wise granularity for weights and layer-wise granularity for activations. 
We used the default setting of 32 sample images for calibration.
}

\subsubsection{\secondR{Settings for FQ-ViT}}
\secondR{Essentially, we performed symmetric quantization on a per-channel basis for weights and asymmetric quantization on a per-layer basis for activations. 
To ensure a fair comparison, we set the quantization for the weights to the minimum and maximum values. The hyperparameter $K$ in the power of two factor remained unchanged. 
For the calibration process, we selected a sample of 1,000 images.}

\subsubsection{\secondR{Settings for Q-HyViT}}
\secondR{We quantized all the weights and inputs for the fully connected layers, including the first projection layer and the last head layer. 
Additionally, the two input matrices for the matrix multiplications in the self-attention modules were quantized. 
The inputs of the softmax and normalization layers were also quantized, which was consistent with FQ-ViT. We used 32 images for calibration, and unoptimized scaling factors were initialized with minimum and maximum values.}

\subsection{Accuracy Evaluation}
We selected MobileViTv1 \cite{mehta2021mobilevit}, MobileViTv2 \cite{mehta2022separable}, Mobile-Former \cite{chen2022mobile}, EfficientFormerV1 \cite{li2022efficientformer}, and EfficientFormerV2~\cite{li2022rethinking} as representative hybrid vision transformer architectures.

\subsubsection{Results with Partial Quantization.}
Table~\ref{table:accuracy} shows quantization results on hybird ViT architectures with varying model sizes, in terms of 8-bit and 6-bit quantization, where softmax and layer-norm remain under floating-point.

Upon analyzing Table~\ref{table:accuracy}, it is clear that prior studies have observed a significant drop in accuracy, even when using 8-bit quantization, in hybrid vision transformers.
PTQ4ViT and EasyQuant perform fairly well by exploring scaling factors for each layer using Hessian and Cosine similarity to minimize reconstruction error when the model size is large.
However, for models with fewer than 5M parameters, such as extremely lightweight models, existing quantization methods inadequately address dynamic activation changes. 
This leads to significant accuracy degradation, even in 8-bit settings. 
\firstR{In the case of RepQ-ViT, accuracy is well preserved when the model size is large, but when the model size is reduced to less than 5M, there is a significant drop in accuracy. 
The RepQ-ViT model lacks the ability to include weight adjustment via a reconstruction approach, hence inadequately addressing the complex activations that occur in hybrid vision transformers. Therefore, it shows lower accuracy compared with prior works that used reconstruction methods (PTQ4ViT and Q-HyViT). Particularly, in Mobile-Former, the granularity adjustment proposed by RepQ-ViT through reparameterization fails to accommodate the changes in activation that occur in hybrid vision transformers, resulting in the most significant drop in accuracy.}

In contrast, our Q-HyViT achieves less than 1\% accuracy drop with 8-bit quantization on the sub-5M models including xxs, xs, 050, 075, 100, 26m, 52m, 96m, and S0.
In summary, Q-HyViT exhibits average improvements of 9.54\% and 7.09\% over EasyQuant and PTQ4ViT, respectively, with an 8-bit setting. Under the 6-bit setup, the improvements reach 43.39\% and 8.65\%, respectively.

%improves on EasyQuant and PTQ4ViT on average by 9.54\% and 7.09\% under an 8-bit setting, respectively, and by 43.39\% and 8.65\% under a 6-bit setting.

\begin{table}[t]
\centering
\caption{\news{Ablation study on hybrid reconstruction error minimization, where \cmark denotes that the component is considered. When all components are disabled, the accuracy results are identical to those of PTQ4ViT.}}
\label{table:ablation}
\resizebox{\columnwidth}{!}{
%\resizebox{5cm}{!}{
\begin{tabular}{ccccc%rr
}
\toprule
\thead{Model \\ Name} &
\thead{Scaling Factor \\ \news{(C3,C4)}} & 
\thead{Granularity \\ \news{(C1,C2,C4)}} & 
\thead{Scheme \\ \news{(C2,C4)}} & 
\thead{Top1 \\ Accuracy} 
\\   \midrule
\midrule
\multirow{4}{*}{MobileViTv1-xxs} & \xmark & \xmark & \xmark & 37.75  \\
  & \cmark & \xmark & \xmark & 44.37  \\
  & \cmark & \cmark & \xmark & 59.50  \\
  & \cmark & \cmark & \cmark & 68.20  \\
\midrule
\multirow{4}{*}{MobileViTv1-xs} & \xmark & \xmark & \xmark &  65.52  \\
  & \cmark & \xmark & \xmark &  69.12 \\
  & \cmark & \cmark & \xmark &  72.00  \\
  & \cmark & \cmark & \cmark & 74.31 \\
\midrule
\multirow{4}{*}{MobileViTv1-s} & \xmark & \xmark & \xmark & 68.19 \\
  & \cmark & \xmark & \xmark & 73.02\\
  & \cmark & \cmark & \xmark & 77.01 \\
  & \cmark & \cmark & \cmark & 77.92 \\
\midrule
\multirow{4}{*}{MobileViTv2-050} & \xmark & \xmark & \xmark & 39.39 \\
  & \cmark & \xmark & \xmark & 49.62\\
  & \cmark & \cmark & \xmark & 69.89 \\
  & \cmark & \cmark & \cmark & 69.89 \\
\midrule
\multirow{4}{*}{MobileViTv2-075} & \xmark & \xmark & \xmark & 65.54 \\
  & \cmark & \xmark & \xmark & 67.24 \\
  & \cmark & \cmark & \xmark & 75.29 \\
  & \cmark & \cmark & \cmark & 75.29 \\
\midrule
\multirow{4}{*}{MobileViTv2-100} & \xmark & \xmark & \xmark & 51.02  \\
  & \cmark & \xmark & \xmark & 68.18\\
  & \cmark & \cmark & \xmark & 77.63 \\
  & \cmark & \cmark & \cmark & 77.63\\
\bottomrule
\multirow{4}{*}{MobileViTv2-125} & \xmark & \xmark & \xmark & 67.39 \\
  & \cmark & \xmark & \xmark & 75.39\\
  & \cmark & \cmark & \xmark & 79.31 \\
  & \cmark & \cmark & \cmark & 79.31\\
\bottomrule
\multirow{4}{*}{MobileViTv2-150} & \xmark & \xmark & \xmark & 68.61 \\
  & \cmark & \xmark & \xmark & 75.88\\
  & \cmark & \cmark & \xmark & 80.09 \\
  & \cmark & \cmark & \cmark & 80.09\\
\bottomrule
\multirow{4}{*}{MobileViTv2-175} & \xmark & \xmark & \xmark & 72.30 \\
  & \cmark & \xmark & \xmark & 76.81\\
  & \cmark & \cmark & \xmark & 80.63 \\
  & \cmark & \cmark & \cmark & 80.63 \\
\bottomrule
\multirow{4}{*}{MobileViTv2-200} & \xmark & \xmark & \xmark & 75.50 \\
  & \cmark & \xmark & \xmark & 77.91\\
  & \cmark & \cmark & \xmark & 80.94 \\
  & \cmark & \cmark & \cmark & 80.94 \\
\bottomrule

\end{tabular}
}

\end{table}

Specifically, MobileViTv2 shows more accuracy drop than MobileViTv1 at the same model size, potentially due to its use of linear attention in self-attention computation, which results in fewer attention maps and lower resilience after post-softmax values.
In the case of EfficientFormerV1, there is no significant difference between the conventional method and the proposed method. The reason for this is that convolution and transformer layers in the meta block are not used in a hybrid manner.
Furthermore, we observe that larger hybrid vision transformers are less sensitive to low-bit quantization (6-bit), as evidenced by the accuracy drops of MobileVitv1-xxs, MobileVitv1-xs, and MobileVitv1-s, which are 2.67\%, 1.36\%, and 1.22\%, respectively. This pattern is also consistent in MobileViTv2, Mobile-Former, EfficientFormerV1, and EfficientFormerV2.
This phenomenon is attributed to larger networks that have more weights and generate more activations, making them more resilient to perturbations caused by quantization.

% result on Fully quantized models
\subsubsection{Results with Full Quantization.}
Previous studies~\cite{liu2021post, yuan2022ptq4vit, wu2020easyquant, apq22} have refrained from quantizing softmax and layer normalization operations due to their smaller computational demand compared to matrix multiplication in terms of total FLOPs. 
Moreover, straightforward quantization of such non-linear functions may result in considerable accuracy degradation. Nonetheless, integer-only quantization~\cite{jacob2018quantization} is important especially for edge and mobile devices. This is due to the fact that softmax operation and layer normalization require dequantization for their computation in floating-point, as well as data movement involved in off-chip memory.
Thus, a fully quantized approach is necessary to alleviate significant hardware design challenges that arise from reducing off-chip level data transfer. 

In line with previous research~\cite{lin2022fq}, we apply a fully quantized approach, FQ-ViT, to hybrid ViTs, as summarized in Table \ref{table:full_accuracy}. Note that FQ-VIT shows very poor accuracy in MobileViTv1-xxs due to its use of an asymmetric scheme with zero points, which fails to handle the high variation of activation range by adjusting quantization granularity. 
\secondR{As the size of the model increases, along with an increase in the number of channels, the effect of zero point overflow on accuracy becomes less significant compared to smaller models.}
%although the same issue arises, as the model size increases, the impact on accuracy diminishes. 
%However, the accuracy drop issue is mitigated as the model size increases.
%In detail, for the FQ-ViT model, the accuracy is 0.1 for the xxs model, but it achieves accuracies of 62.2 and 74.94 for the xs and s models, respectively. 
%In other words, 
%The reason for this mitigation is that the number of redundant parameters increases as the size of the model increases. This is similar to why pruning works well for large models with many redundant parameters such as ResNet.

Furthermore, in the case of MobileViTv2, group normalization is utilized instead of batch and layer norms, causing the existing L2 norm-based scaling factor exploration to function inaccurately. 
Our study addresses these issues and achieves an average of 43.63\% accuracy improvement over FQ-ViT.

\subsection{\secondR{Impact of Calibration Sample Size on Model Accuracy}}
\secondR{
As listed in Table~\ref{table:accuracy} (partial quantization) and Table~\ref{table:full_accuracy} (full quantization), the accuracy of each model varies depending on whether the softmax and layer-norm are quantized and the number of images utilized during the calibration phase. For a fair comparison with EasyQuant and PTQ4ViT in Table~\ref{table:accuracy}, 32 images were used to obtain the results. 
In the case of FQ-ViT, the calibration utilized the 1,000 images referenced in the original paper. 
When increasing the number of images for calibration in partial quantization from 32 to 128, there are no significant differences in accuracy, as listed in Table~\ref{table:numberofimages}.
}

\begin{table}[h]
\centering
\caption{Quantization results under different numbers of calibration images. Partial means that \texttt{softmax} and \texttt{layer-norm} remain in floating-point, while full means that their operators are quantized}
\label{table:numberofimages}
%\resizebox{\columnwidth}{!}{
\begin{tabular}{lcccccccc}
\toprule
\multicolumn{1}{c}{\textbf{Model}} &
\multicolumn{1}{c}{\textbf{Quant.}} &
\multicolumn{1}{c}{\textbf{\# of images}} &
\multicolumn{1}{c}{\textbf{Accuracy}} 
\\   \midrule
\multirow{4}{*}{MobileViTv1-xxs} & Partial & 32 & 68.20  \\
  & Partial & 128 & 68.18   \\
  & Full & 500 & 67.16   \\
  & Full & 1,000 & 67.20   \\
\midrule
\multirow{4}{*}{MobileViTv1-xs} & Partial & 32 & 74.31  \\
  & Partial & 128 & 74.25    \\
  & Full & 500 & 73.82   \\
  & Full & 1,000 & 73.89   \\
\midrule
\multirow{4}{*}{MobileViTv2-s} & Partial & 32 & 77.92  \\
  & Partial & 128 & 77.67   \\
  & Full & 500 & 77.69   \\
  & Full & 1,000 & 77.72   \\
\midrule
\multirow{4}{*}{MobileViTv2-050} & Partial & 32 & 69.89  \\
  & Partial & 128 & 69.89 \\
  & Full & 500 & 68.52  \\
  & Full & 1,000 & 68.73  \\
\midrule
% ours
\multirow{4}{*}{MobileViTv2-075} & Partial & 32 &  75.29 \\
  & Partial & 128 & 75.32 \\
  & Full & 500 & 74.26\\
  & Full & 1,000 & 74.36  \\
% ours
\midrule
\multirow{4}{*}{MobileViTv2-100} & Partial & 32 &  77.63 \\
  & Partial & 128 & 77.75 \\
  & Full & 500 & 77.19 \\
  & Full & 1,000 & 77.13  \\
\bottomrule
\end{tabular}
%}
\end{table}

\subsection{\secondR{Running time of quantization methods}}
\begin{figure}[t]
	\centering
	\includegraphics[width=0.8\columnwidth]{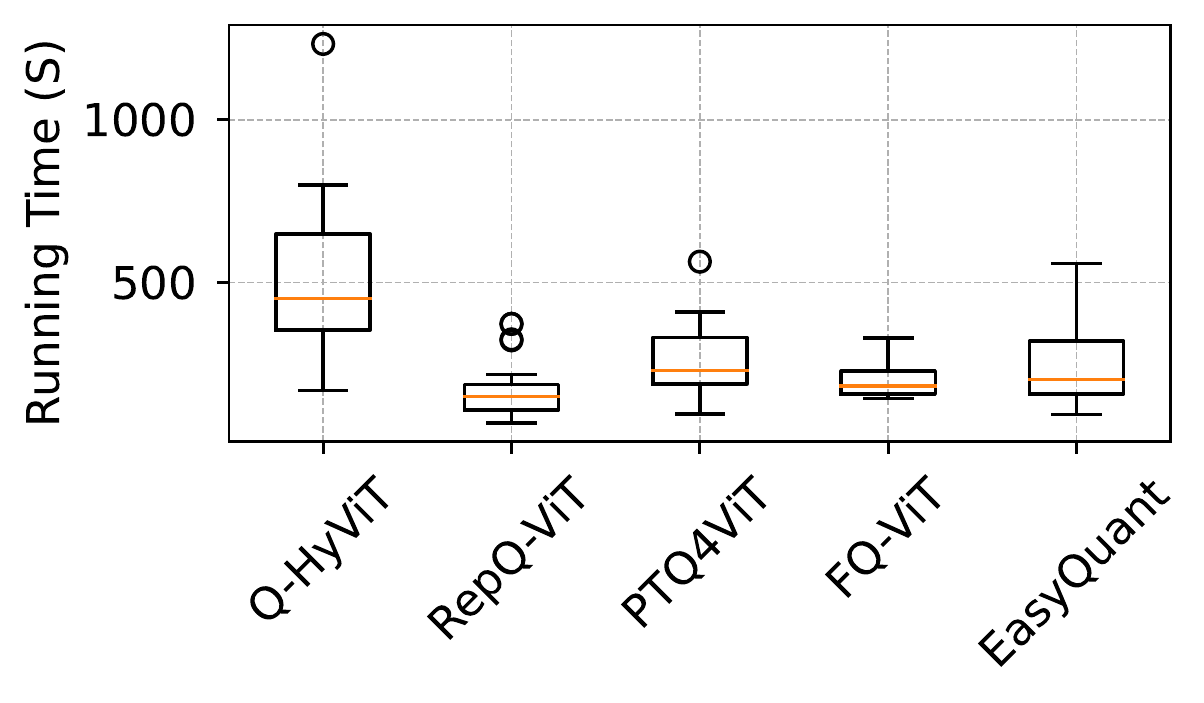}
	\caption{\secondR{Running time of quantization methods}}
	\label{fig:running_time}
\end{figure}

\secondR{
To compare the running times among various quantization methods, we measured the running times for five models using one A100-80G GPU for each quantization method.
The results are shown in Figure~\ref{fig:running_time}, where the proposed Q-HyViT takes the longest with an average of 523 seconds, and the RepQ-ViT method takes the shortest time with an average of 159 seconds. 
These differences are influenced by whether the method performs reconstruction and the number of images used for calibration during quantization.
Reconstruction methods, such as EasyQuant, PTQ4ViT, and the proposed Q-HyViT, have a longer running time than methods that do not perform reconstruction. 
Both FQ-ViT and RepQ-ViT do not use reconstruction; however, FQ-ViT requires 1,000 images for calibration, thus consuming more time than RepQ-ViT, which uses only 32 images.
However, these differences are trivial when compared to the substantial number of GPU hours produced by quantization-aware training methods. 
Since post-training quantization is performed only once offline, these minute differences are negligible.
}

\subsection{Ablation Study on Hybrid Reconstruction}
We performed an ablation study to assess the impact of utilizing the optimal selection of scaling factors, granularity, and quantization scheme within the context of hybrid reconstruction error minimization. 
\news{Furthermore, every functional component in hybrid vision transformers tackles one of the four challenges that arise. Conducting an ablation study enables us to assess the degree of mitigation empirically.}

Table \ref{table:ablation} provides a summary of the ablation study results for various sizes and types of hybrid vision transformer architectures, where the results demonstrate that all the components enhance the accuracy of quantized hybrid ViTs.

When only optimizing the scaling factor with hybrid reconstruction error minimization, it does not yield significant performance improvements compared to PTQ4ViT~(baseline).
However, optimizing not only the scaling factors but also the granularity and scheme based on the bridge block leads to synergistic effects in general.
As a result, combining all of them together achieves significant accuracy improvements.
In contrast, excluding the optimization of granularity significantly decreases the accuracy, highlighting highly dynamic activation ranges in lightweight hybrid ViTs as the main cause of the accuracy drop.

\section{\secondR{Implications and Recommendations for IoT}}
\secondR{
Quantization method significantly contributes to the IoT domain by enabling efficient computation and communication, which are essential for deploying artificial intelligence and machine learning models on IoT devices. 
We describe a roadmap for integrating quantization in IoT applications, system software support, and hardware design.
}

\secondR{
\textbf{Enhancing IoT Applications through Efficient Federated Learning}
FedQNN~\cite{ji2022fedqnn} and QuAsyncFL~\cite{liu2023quasyncfl} showcase how quantization facilitates federated learning in IoT by reducing the model's bitwidth, thus lowering the computational and communication overhead. 
This enables IoT devices to participate in federated learning networks more effectively, allowing for distributed, privacy-preserving machine learning without the need for high bandwidth or powerful computational resources. By adopting low-bitwidth neural network quantization and asynchronous federated learning approaches, IoT applications can achieve smarter data processing and decision-making capabilities, improving areas such as smart agriculture, healthcare monitoring, and urban traffic management.}

\secondR{
{\bf IoT Hardware Design support:}}
\secondR{The implementation of efficient neural networks within the IoT domain necessitates a reevaluation of hardware design for IoT devices. For this reason, hardware tailored for efficient neural network inference on IoT devices has been developed~\cite{yang2023parallel,russo2021dnn}.
These specialized hardware solutions for the IoT aim to reduce design complexity and enhance power efficiency by supporting 8-bit operations. Achieving this operational support requires the use of quantization techniques, which are essential and can also leverage the quantization methods proposed in this research.}

%------------------------------------------------------------------------
\section{Conclusion}

We addressed the problem of democratizing vision transformers on resource-constrained devices by proposing a method for minimizing quantization errors in hybrid vision transformers. 
The proposed method, Q-HyViT, identified the four challenges of applying post-training quantization (PTQ) to hybrid vision transformers and proposed a unified method to mitigate errors in PTQ. 
Q-HyViT achieved this by selecting optimal scale factors, granularity, and scheme for both bridge and non-bridge layers based on hybrid reconstruction error minimization from a loss degradation perspective.
We demonstrated the effectiveness of Q-HyViT by conducting extensive experiments comparing it with existing several open-source algorithms, EasyQuant, FQ-ViT, PTQ4ViT, and RepQ-ViT on the same hybrid vision transformers. 
The results demonstrated that Q-HyViT outperforms existing methods by a significant margin and achieved state-of-the-art accuracy on hybrid vision transformers in a fully quantized manner, including non-linear operations such as softmax and diverse normalization.
Finally, we contributed to the field of artificial intelligence by identifying the four unique challenges of quantizing hybrid vision transformers and proposing an effective solution for minimizing quantization error. 
%Our work has significant implications for democratizing vision transformers on resource-constrained devices, making it possible to use these models in real-world applications where computational resources are limited.

%\newpage

% \section*{Acknowledgments}
% This should be a simple paragraph before the References to thank those individuals and institutions who have supported your work on this article.


\iffalse
\begin{figure}[t]
\centering
\subfloat[][]{\label{sub:hist_channel1_quant}\includegraphics[width=.4\textwidth]{./figures/59_channelwise}} \vfill
\subfloat[][]{\label{sub:hist_channel5_quant}\includegraphics[width=.4\textwidth]{./figures/59_layerwise}}
\caption{An overlapping histogram of quantized values (blue) and real values (orange) in the activation of  for 1st bridge block of MobileViTv1-xxs: (left) layer-wise quantization (right) channel-wise quantization}
\label{fig:channel_quant}
\end{figure}
\fi

%\section{Overflow Issues in Asymmetric Channel-wise Quantization}

%}

%\nolinenumbers

\bibliographystyle{IEEEtran}
\bibliography{iot}

%\newpage

\begin{IEEEbiography}[{\includegraphics[width=1in,height=1.25in,clip,keepaspectratio]{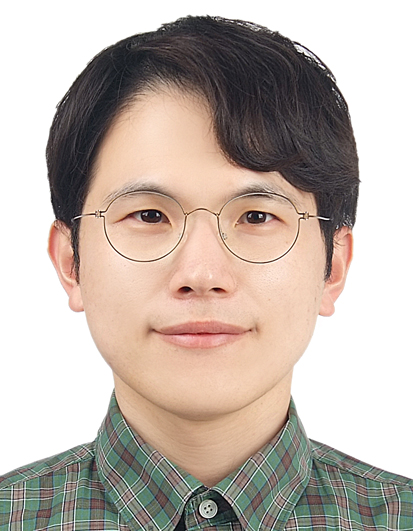}}]{Jemin Lee} received his B.S. and Ph.D. degrees in computer science and engineering from Chungnam National University in 2011 and 2017, respectively. He is currently a senior researcher at the Electronics and Telecommunications Research Institute (ETRI). Since 2023, he has also served as an assistant professor in the AI Department at the University of Science and Technology (UST). Previously, he was a postdoctoral researcher at the Korea Advanced Institute of Science and Technology (KAIST) from 2017 to 2018. His research interests include energy-aware mobile computing and deep learning compilers.
%received the B.S. and Ph.D. degrees in computer science and engineering from Chungnam National University in 2011 and 2017, respectively. He is a senior researcher at the Electronics and Communications Research Institute (ETRI). Since 2023, he has also been with the AI Department, the University of
%Science and Technology (UST), where he is currently an assistant professor. He was a post-doctoral researcher at the Korea Advanced Institute of Science and Technology (KAIST) in 2017–2018. His research interests include energy-aware mobile computing and deep learning compiler.
\end{IEEEbiography}

\begin{IEEEbiography}
[{\includegraphics[width=1in,height=1.25in,clip,keepaspectratio]{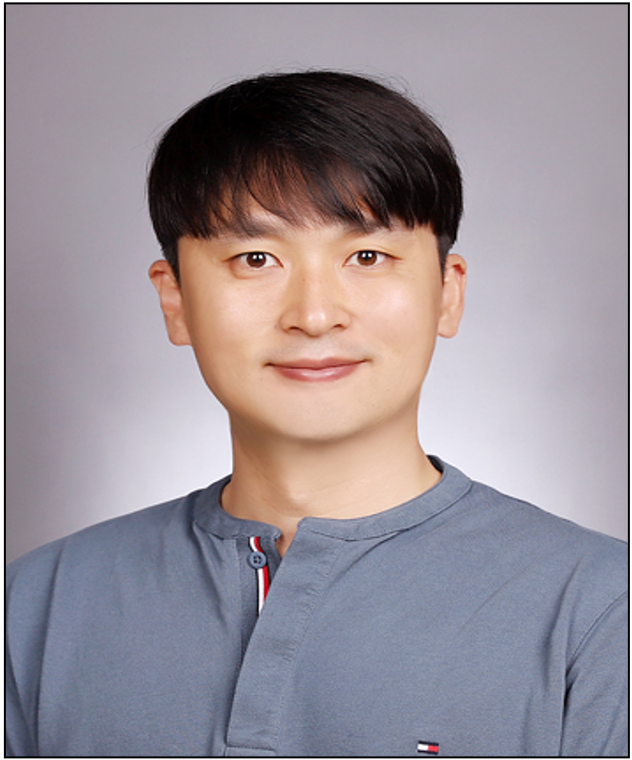}}]{Yongin Kwon} received the B.Sc. degree in electrical and electronic engineering from the Korea Advanced Institute of Science and Technology, South Korea, in 2008, and M.S. and Ph.D. degrees in electrical and computer engineering from Seoul National University, South Korea, in 2010 and 2015, respectively. From 2015 to 2019, he worked at Samsung Electronics as a Staff Software Engineer. He has been with Electronics and Telecommunications Research Institute (ETRI) since 2019, where he is currently a Senior Researcher. His research interests include neural processing units, compiler, deep learning, and embedded systems.
\end{IEEEbiography}

\begin{IEEEbiography}
[{\includegraphics[width=1in,height=1.25in,clip,keepaspectratio]{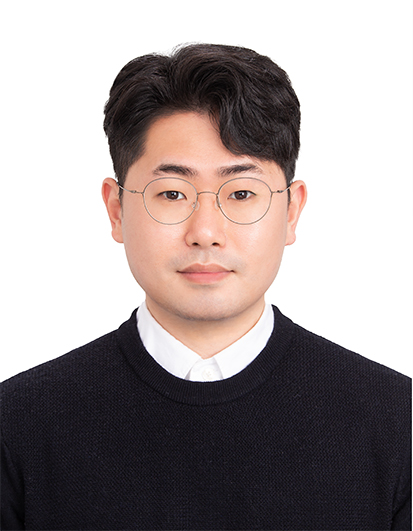}}]{Sihyeong Park} received the B.S., M.S., and Ph.D. degrees in computer science and engineering from Chungnam National University, in 2014, 2016, and 2021, respectively. He is a senior researcher at the Korea Electronics Technology Institute (KETI). His research interests include multi-core embedded systems and real-time systems.
\end{IEEEbiography}

\begin{IEEEbiography}
[{\includegraphics[width=1in,height=1.25in,clip,keepaspectratio]{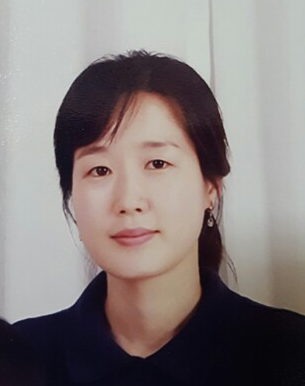}}]{Misun Yu} received the M.S. degree from the Department of Computer Science and Engineering at Pohang University of Science and Technology, Republic of Korea.
She is a principal researcher at the Electronics and Communications Research Institute (ETRI), Daejeon, Rep. of Korea. Her main research interests include concurrent program analysis, software testing, deep learning, and embedded systems.
\end{IEEEbiography}

\begin{IEEEbiography}
[{\includegraphics[width=1in,height=1.25in,clip,keepaspectratio]{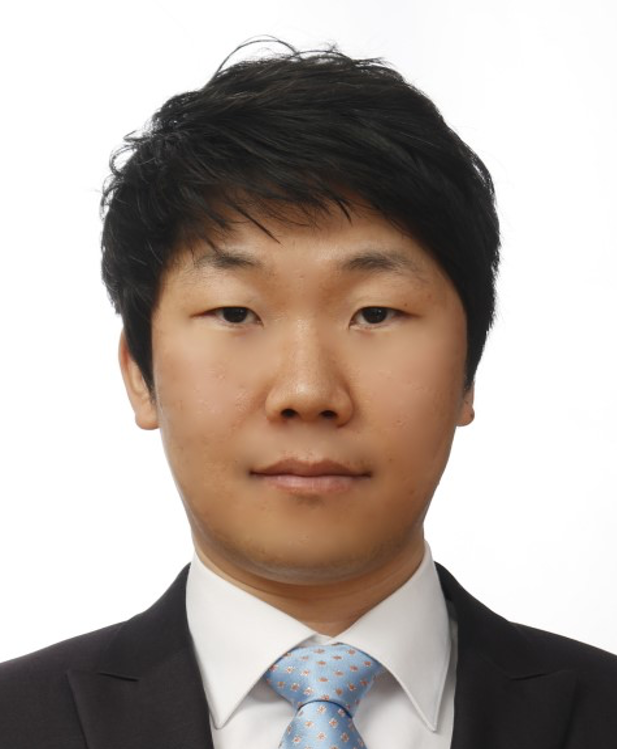}}]{Jeman Park} received his B.S., M.S., and Ph.D. degrees in electronics and computer engineering in Hanyang University, Republic of Korea, in 2004, 2006, and 2014, respectively. Since 2019, he has been with Electronics and Telecommunications Research Institute, Daejeon, Republic of Korea. where he is now a senior researcher. His main research interests are computer network, edge computing, and AI compiler.
\end{IEEEbiography}

\begin{IEEEbiography}
[{\includegraphics[width=1in,height=1.25in,clip,keepaspectratio]{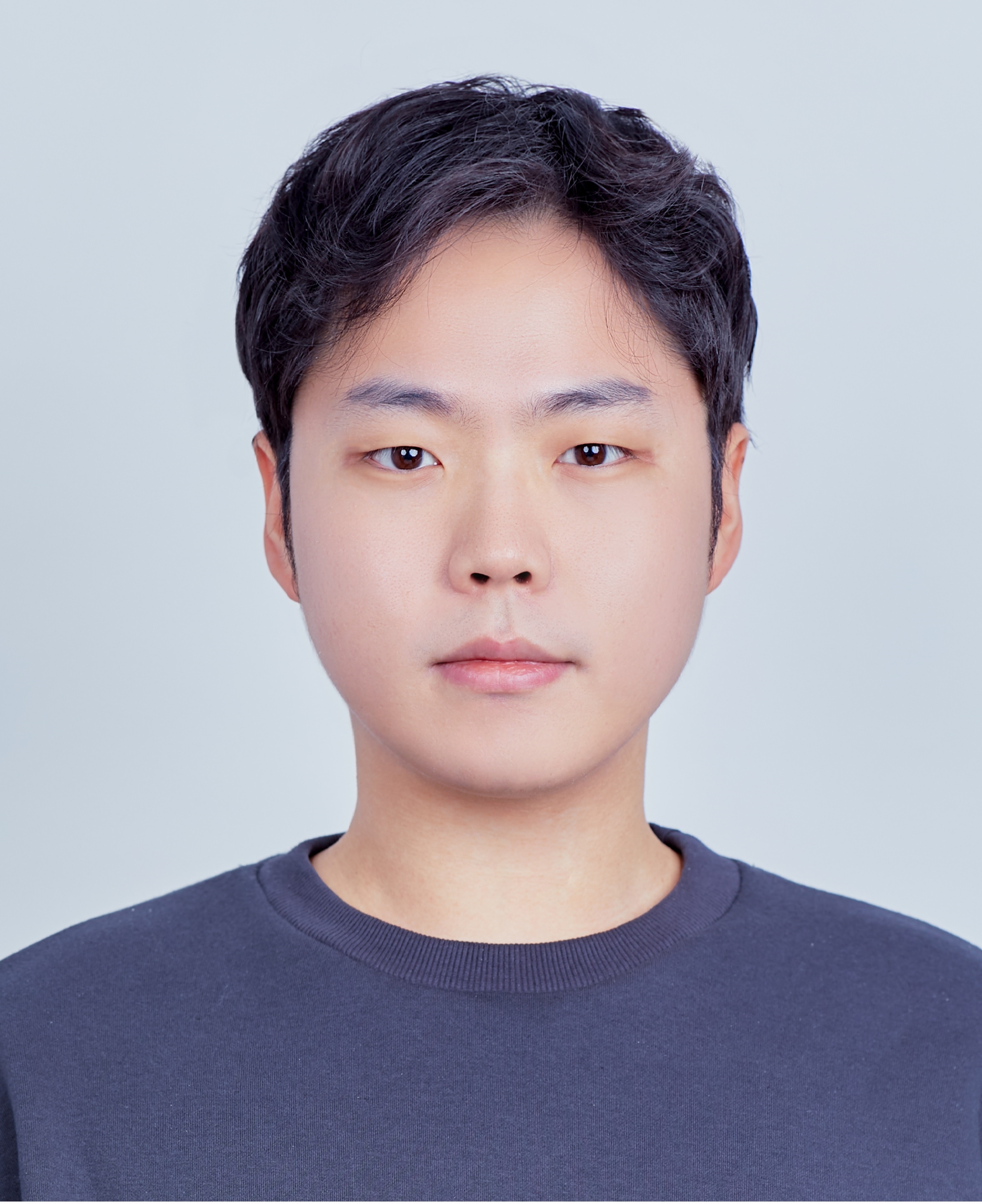}}]{Hwanjun Song} is an assistant professor in the Department of Industrial and Systems Engineering at KAIST. Previsouly, he was a Research Scientist at AWS AI Labs in 2023 and at NAVER AI Lab in 2021--2022, and Research Intern
at Google Research in 2020. He earned his
Ph.D. degree in the Graduate School of Data
Science from KAIST in 2021.
He is interested in designing advanced methodologies to handle data scale and quality issues,
which are two main real-world challenges for AI.
He was sponsored by Microsoft through Azure
for Research from 2016 to 2018, and received
the Qualcomm Innovation Award in 2019.
\end{IEEEbiography}

\vfill

\end{document}